\documentclass{article}
\usepackage{float}
\usepackage{PRIMEarxiv}
\usepackage{multirow}
\usepackage[numbers]{natbib}
\usepackage[utf8]{inputenc}
\usepackage[T1]{fontenc} 
\usepackage{arydshln}
\usepackage[colorlinks=true,citecolor=blue]{hyperref}
\usepackage{amsmath}
\usepackage{url}
\usepackage{booktabs}
\usepackage{enumitem}
\usepackage{amsfonts}
\usepackage{nicefrac}
\usepackage{microtype}
\usepackage{lipsum}
\usepackage{fancyhdr}
\usepackage{graphicx}
\graphicspath{{media/}}
\usepackage[table]{xcolor}
\DeclareGraphicsExtensions{.pdf,.png,.jpg}

\usepackage{fontawesome5,xcolor}
\definecolor{gold}{HTML}{C9A227}
\definecolor{silver}{HTML}{B4B4B4}
\definecolor{bronze}{HTML}{B87333}

\newcommand{\trophyscale}{0.68}
\newcommand{\awardscale}{0.81}
\newcommand{\Gold}{\raisebox{0.1ex}{\scalebox{\trophyscale}{\textcolor{gold}{\faTrophy}}}}
\newcommand{\Silver}{\raisebox{0.1ex}{\scalebox{\awardscale}{\textcolor{silver}{\faAward}}}}
\newcommand{\Bronze}{\raisebox{0.1ex}{\scalebox{\awardscale}{\textcolor{bronze}{\faAward}}}}

\title{TANTE: Time-Adaptive Operator Learning via Neural Taylor Expansion}

\begin{document}

\author{ 
Zhikai Wu\thanks{Work completed as an intern at Yale University.} \ \ \
Sifan Wang \ \
Shiyang Zhang \ \
Sizhuang He \ \
Min Zhu \ \
Anran Jiao \\ 
\textbf{Lu Lu}\thanks{Corresponding authors. Email: \texttt{lu.lu@yale.edu},$\,$ \texttt{david.vandijk@yale.edu}.} \quad 
\textbf{David van Dijk}\footnotemark[2]\\
[4pt] 
Yale University, New Haven, CT, USA\\
}

\maketitle

\begin{abstract}
Operator learning for time-dependent partial differential equations (PDEs) has seen rapid progress in recent years, enabling efficient approximation of complex spatiotemporal dynamics. However, most existing methods rely on fixed time step sizes during rollout, which limits their ability to adapt to varying temporal complexity and often leads to error accumulation. Here, we propose the Time-Adaptive Transformer with Neural Taylor Expansion (TANTE), a novel operator-learning framework that produces continuous-time predictions with adaptive step sizes. TANTE predicts future states by performing a Taylor expansion at the current state, where neural networks learn both the higher-order temporal derivatives and the local radius of convergence. This allows the model to dynamically adjust its rollout based on the local behavior of the solution, thereby reducing cumulative error and improving computational efficiency. We demonstrate the effectiveness of TANTE across a wide range of PDE benchmarks, achieving superior accuracy and adaptability compared to fixed-step baselines, delivering accuracy gains of 60--80\% and speed‑ups of 30--40\% at inference time. The code is publicly available at \href{https://github.com/zwu88/TANTE}{\texttt{https://github.com/zwu88/TANTE}} for transparency and reproducibility.
\end{abstract}


\section{Introduction}

Operator learning seeks to approximate mappings between infinite-dimensional function spaces, offering a data-driven alternative to traditional numerical solvers for partial differential equations (PDEs)~\cite{lu2021learning,karniadakis2021physics,jiao2025one}. This approach has gained increasing attention for its potential to model complex physical systems efficiently, especially when classical solvers are computationally prohibitive. By directly learning the solution operator from data, these methods enable generalization across varying initial conditions and mesh resolutions—properties essential for various scientific and engineering applications, such as weather forecasting \cite{pathak2022fourcastnet,lam2023learning}, industrial-scale automotive aerodynamics \cite{li2023geometryinformedneuraloperatorlargescale}, dynamic urban microclimate \cite{peng2024fourier}, material deformation \cite{liu2022learning,rashid2022learning}, mechanics \cite{wang2024micrometer,kou2025neural}, lithography \cite{yang2022generic}, geophysics~\cite{lee2024efficient}, photoacoustics \cite{dreier2019operator}, and electromagnetic fields \cite{peng2022rapid}.

\begin{figure}[htbp]
\begin{center}
\includegraphics[width=\textwidth]{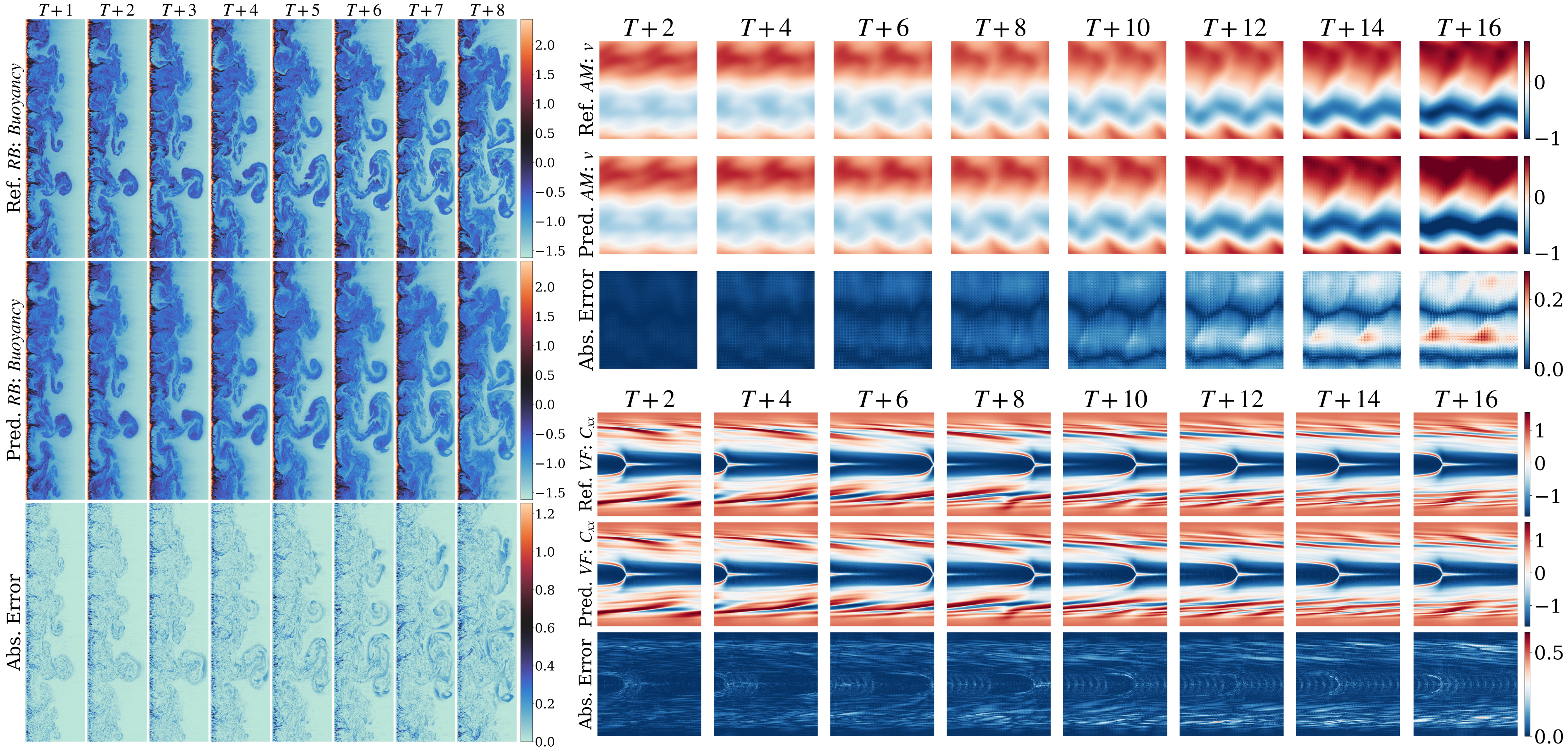} \\
\includegraphics[width=\textwidth]{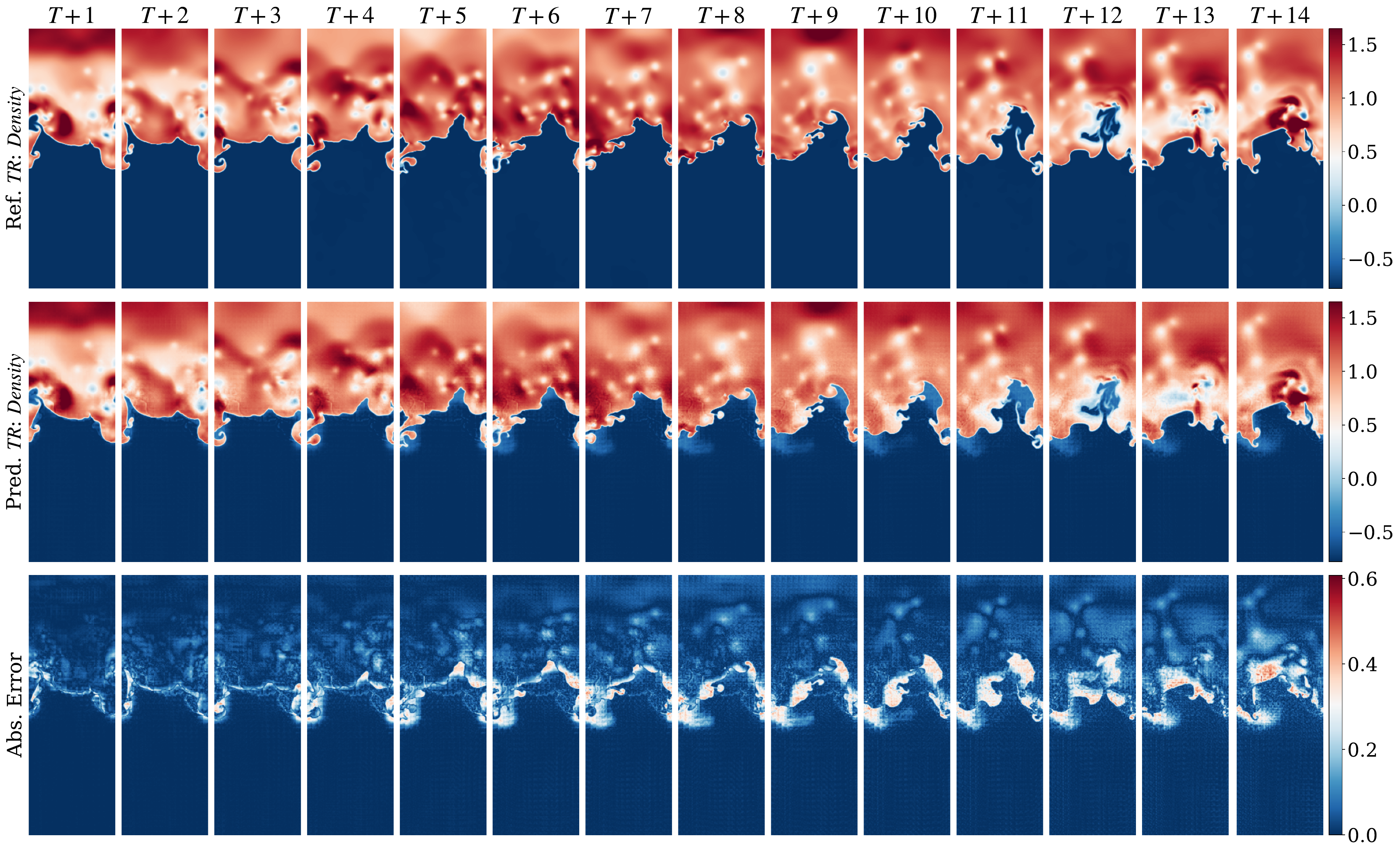}
\caption{\textit{Representative TANTE rollout predictions across four benchmarks.} Each benchmark's results are shown in three rows: the first row displays the ground truth field (reference), the second row shows the predictions from TANTE, and the third row illustrates the point-wise absolute error between the predictions and the ground truth. 
\textbf{Top Left:} $Buoyancy$ field in the \textit{Rayleigh-Bénard Convection (RB)} benchmark across eight time steps. 
\textbf{Top Right:} $Velocity$ field (y-direction) in the \textit{Active Matter (AM)} benchmark across sixteen time steps. 
\textbf{Middle Right:} $C_{xx}$ field in the \textit{Viscoelastic Fluids (VF)} benchmark across sixteen time steps.
\textbf{Bottom:} $Density$ field in the \textit{Turbulent Radiative Layer (TR)} benchmark across fourteen time steps. 
}
\label{main_viz_3}
\end{center}
\end{figure}

Recent work has demonstrated the feasibility of operator learning in both spatial and spatiotemporal domains. Foundational architectures, such as Deep Operator Network (DeepONet)~\cite{lu2021learning,lu2022comprehensive} and the Fourier Neural Operator (FNO)~\cite{li2020fourier} introduced end-to-end frameworks for approximating nonlinear operators, inspiring a series of architectural extensions. Variants including graph neural operator~\cite{NEURIPS2020_4b21cf96}, multiple-input DeepONet~\cite{jin2022mionet}, multifidelity DeepONet~\cite{lu2022multifidelity}, U-Fourier neural operator~\cite{WEN2022104180}, wavelet neural operator~\cite{tripura2022waveletneuraloperatorneural}, Adaptive Fourier Neural Operator (AFNO)~\cite{guibas2022adaptivefourierneuraloperators}, convolutional neural operator~\cite{raonić2023convolutionalneuraloperatorsrobust}, Laplace neural operator~\cite{cao2023lnolaplaceneuraloperator}, spectral neural operator~\cite{fanaskov2024spectralneuraloperators}, U-shaped Neural Operator (UNO)~\cite{rahman2023uno}, and quantum DeepONet~\cite{xiao2025quantum}, have enhanced this paradigm by improving expressivity, scalability, and generalization across diverse PDE types. Transformer-based neural operators—such as Galerkin Transformer~\cite{cao2021choosetransformerfouriergalerkin}, operator Transformer~\cite{li2023transformerpartialdifferentialequations}, factorized Transformer~\cite{li2023scalabletransformerpdesurrogate}, vision Transformer-operator~\cite{ovadia2023vitovisiontransformeroperator}, Continuous Vision Transformer (CViT)~\cite{wang2025cvitcontinuousvisiontransformer}, and diffusion models over function spaces~\cite{wang2025fundiff}---further advance this line of work by leveraging self-attention to capture global dependencies and long-range spatiotemporal dynamics more effectively.

Foundation models for PDEs—such as Denoising Pre-training Operator Transformer (DPOT)~\cite{hao2024dpot}, Poseidon~\cite{herde2024poseidonefficientfoundationmodels}, multiple physics pretraining~\cite{NEURIPS2024_d7cb9db5}, and PROSE-FD~\cite{liu2024prosefdmultimodalpdefoundation}—seek to generalize across a wide range of operators and boundary conditions via large-scale pretraining. In parallel, to address PDEs defined on irregular domains or complex geometries, models like diffeomorphic mapping operator learning~\cite{yin2024scalable}, Transolver~\cite{wu2024transolverfasttransformersolver}, universal physics Transformer~\cite{alkin2025universalphysicstransformersframework}, general neural operator Transformer~\cite{hao2023gnotgeneralneuraloperator}, and geometry-informed neural operator~\cite{li2023geometryinformedneuraloperatorlargescale} incorporate geometric priors or mesh-based representations. Another complementary direction integrates physics-based inductive biases into operator learning: hybrid methods like physics-informed neural operator~\cite{wang2021learning,li2023physicsinformedneuraloperatorlearning,ouyang2025rams}, PDEformer~\cite{ye2025pdeformerfoundationmodelonedimensional}, and neural-operator element method~\cite{ouyang2025neural} embed governing physical laws through physics-informed objectives, improving accuracy and data efficiency in low-data or extrapolative regimes. 

For time-dependent PDEs in particular, most of these methods adopt an autoregressive rollout strategy, where predictions are made sequentially over time using a fixed step size. Temporal information is typically handled either by concatenating time as an additional input feature or by modeling time as a sequential axis over which dynamics are learned. Although effective, these designs depend on uniform time discretization, which limits their ability to adapt to changes in temporal complexity during evolution. In reality, physical phenomena often evolve unevenly over time, with periods of gradual change interrupted by sudden transitions. Fixed time-step rollouts fail to account for this structure, resulting in inefficient computations and greater error accumulation in regions where changes occur rapidly.  To address this, we propose the Time-Adaptive Transformer with Neural Taylor Expansion (TANTE), a novel framework for continuous-time operator learning. Our contributions are summarized as follows.
\begin{itemize}[leftmargin=*]
\item We introduce the Neural Taylor Expansion, a novel continuous-time prediction formulation, where neural networks jointly learn high-order temporal derivatives and the local radius of convergence, enabling evaluation at arbitrary time points within the confidence interval.
\item To realize this formulation, we propose a transformer-based architecture that adaptively adjusts temporal resolution based on the local complexity of the solution, allowing for accurate and efficient rollout over uneven dynamics.
\item We validate our approach on a suite of challenging PDE benchmarks, where TANTE achieves 60--80\% improvements in predictive accuracy and 30--40\% reductions in inference time compared to fixed-step baselines. Figure~\ref{main_viz_3} provides representative rollouts predicted by TANTE.
\end{itemize}
Our approach represents a significant shift toward temporal adaptivity in neural operators, opening new possibilities for modeling multi-scale phenomena where computational resources can be dynamically allocated to time regions of greatest complexity. Beyond immediate performance gains, TANTE establishes a foundation for future research at the intersection of scientific machine learning and adaptive numerical methods. 

\section{Methods}
\label{Method}

\begin{figure}[htbp]
\begin{center}
\centerline{\includegraphics[width=\textwidth]{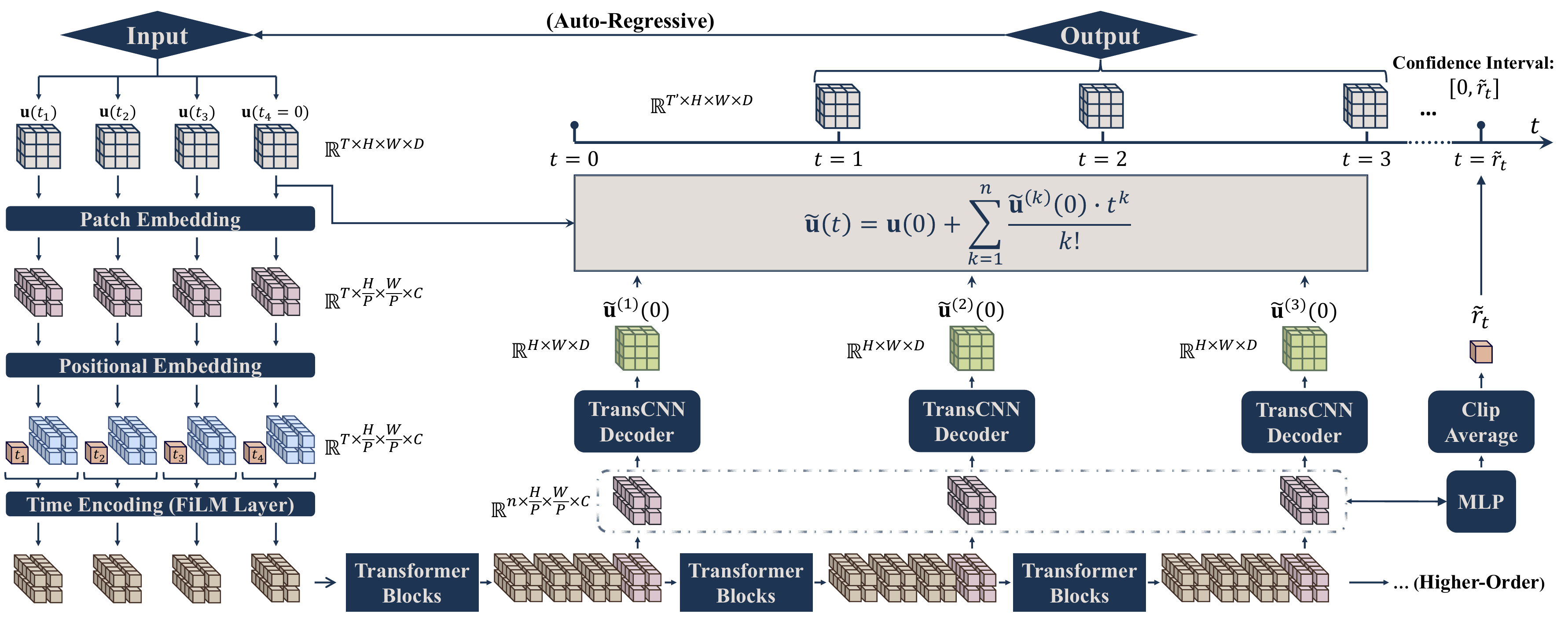}}
\caption{\emph{Time-Adaptive Transformer with Neural Taylor Expansion (TANTE)}. Our framework enables continuous-time prediction with dynamically adjusted step sizes based on the local temporal complexity. TANTE consists of three main components:
(a) a spatiotemporal encoder that extracts spatial tokens from input frames and modulates them with temporal information via a FiLM layer; 
(b) a Transformer Processor that estimates multi-order temporal derivatives at the most recent timestamp $t=0$, with each group of blocks predicting one derivative order; 
(c) a spatiotemporal decoder that predicts derivatives and infers a confidence interval $[0, \tilde{r}_t]$, defining the time range within which the Taylor expansion is valid.
TANTE generates forecasts by summing the predicted derivatives as a Taylor series within the confidence interval.
When predictions extend beyond the confidence interval, the model operates autoregressively, incorporating previously predicted states into the next-step input sequence.
}
\label{pipeline}
\end{center}
\vspace{-5mm}
\end{figure}

Conventional operator learning methods usually use uniform time discretization, which fails to adapt to the uneven temporal complexity inherent in many physical systems. This inefficiency leads to unnecessary computations during periods of gradual change and insufficient resolution during rapid transitions.
To overcome these limitations, we propose Time-Adaptive Transformer with Neural Taylor Expansion (TANTE), a framework that enables temporally continuous prediction with adaptive time stepping informed by the underlying dynamics (Figure~\ref{pipeline}). TANTE achieves this by learning not only high-order temporal derivatives but also estimating the local time interval within which a Taylor series expansion remains valid. This allows the model to dynamically adjust prediction step sizes based on the local dynamical behavior. The architecture consists of three key components: (a) a spatiotemporal encoder that processes sequential spatial inputs at continuous time points, (b) a Transformer Processor with attention along spatiotemporal dimensions, and (c) a decoder that estimates multi-order derivatives and infers a confidence interval for rollout. By summing the predicted derivatives as a Taylor series, TANTE generates continuous-time forecasts via adaptive rollout according to local temporal complexity.

\subsection{Spatiotemporal encoder}
\label{Spatiotemporal-Encoder}
The TANTE encoder processes a spatiotemporal input tensor $\mathbf{u} \in \mathbb{R}^{T \times H \times W \times D}$, representing a sequence of $T$ spatial frames $\mathbf{S}_i \in \mathbb{R}^{H \times W \times D}$ with $D$ physical channels. It also takes a corresponding sequence of relative timestamps $\mathbf{T}_{\text{seq}} \in \mathbb{R}^T$, which represent the time distances between each spatial frame and the most recent frame.

We define the most recent frame $\mathbf{S}_T$ to be at $t = 0$, such that $\mathbf{u}(0) = \mathbf{S}_T$. Each spatial frame is independently tokenized using a convolutional neural network, resulting in patchified inputs $\mathbf{u}_p \in \mathbb{R}^{T \times (H/P) \times (W/P) \times C}$, where $P$ is the patch size and $C$ is the embedding dimension. Let $H'=H/P$ and $W'=W/P$, so that $\mathbf{u}_p \in \mathbb{R}^{T \times H' \times W' \times C}$.

We add learnable 2D spatial positional embeddings $\mathbf{PE}_s$ and modulate each frame using Feature-wise Linear Modulation layer (FiLM)~\cite{perez2017filmvisualreasoninggeneral} to embed its corresponding temporal information:
\begin{equation*}
\mathbf{u}_s = \mathbf{u}_p + \mathbf{PE}_s,\ \mathbf{PE}_s \in \mathbb{R}^{1\times H'\times W'\times C};\ \ \ \mathbf{u}_{st} = \mathrm{FiLM}(\mathbf{T}_{seq}, \mathbf{u}_s).
\end{equation*}

Here, the FiLM layer generates two learnable transformations $\gamma(t)$ and $\beta(t)$ based on the input timestamp $t\in\mathbf{T}_{\mathrm{seq}}$. The spatial feature $\mathbf{u}_{s}(t)$ is then modulated as
\begin{equation*}
    \mathbf{u}_{st}(t) = \gamma(t)\odot \mathbf{u}_{s}(t)+\beta(t),
\end{equation*}
where $\odot$ represents element-wise multiplication. This design allows a single FiLM module to condition $\mathbf{u}_s$ using a time-varying feature $\mathbf{T}_{\mathrm{seq}}$.

\subsection{Transformer processor} 
\label{Transformer-Processor}
We use $n$ groups of Transformer blocks to estimate the first $n$ temporal derivatives at $t=0$, with each block having an embedding size of $C$. We employ three types of Transformer blocks, each with attention mechanisms along different spatiotemporal dimensions: the time dimension ($T$-axis), the height dimension ($H'$-axis), and the width dimension ($W'$-axis). We stack these three types of Transformer blocks crosswise to extract features along each axis. 

To obtain the latent representation of the $k$-th order derivative at $t=0$, we extract the tokens at the most recent timestamp: $\mathbf{z}_k \in \mathbb{R}^{1\times H'\times W'\times C}$ from the output of the $k$-th group of Transformer blocks, resulting in $\mathbf{z} \in \mathbb{R}^{n \times 1\times H' \times W' \times C}$ for all derivatives.
We denote by $\mathbf{z}_k$ the representation corresponding to the estimated local $k$-th order derivative at time $t=0$. 

\subsection{Spatiotemporal decoder}
\label{Spatiotemporal-Decoder}
The decoder performs two key functions: determining the prediction confidence interval and reconstructing spatial derivatives. For the confidence interval, we apply one MLP to all tokens in the latent representation $\mathbf{z}_k$ of each order and average the results to obtain the time radius of convergence:
\begin{equation*}
    \tilde{r}_{t}=\frac{1}{n\cdot H'\cdot W'}\sum_{k=1}^n\sum_{i,\ j}\mathrm{MLP}_k(\mathbf{z}_{k,\ i,\ j}),\quad
    \begin{array}{l}
    i\, \in \{1, 2, \dots, H'\}, \\
    j \in \{1, 2, \dots, W'\}.
    \end{array}
\end{equation*}
This radius $\tilde{r}_{t}$ defines our confidence interval $[0,\ \tilde{r}_t]$. To avoid a degenerate estimate of the interval, we constrain $\tilde{r}_t$ with upper and lower bounds and introduce a regularization loss term penalizing overly small values. We use a piecewise power-exponential function as our regularization loss term:
\begin{equation*}
\mathcal{L}_{r}(\tilde{r}_t) =
\left \{
\begin{array}{cl}
(1+\varepsilon-\tilde{r}_t)^m, & \textrm{if}\ \tilde{r}_t\le1+\varepsilon, \\
0, & \textrm{if}\ \tilde{r}_t>1+\varepsilon.
\end{array}
\right.
\end{equation*}
A detailed ablation study of the hyperparameters $\varepsilon$ and $m$ is provided in Section~\ref{Ablation}.

To obtain each $k$th-order predicted derivative $\tilde{\mathbf{u}}^{(k)}(0) \in \mathbb{R}^{H\times W\times D}$ from its corresponding representation $\mathbf{z}_k$, we employ dedicated Transposed convolutional neural networks (TransCNN):
\begin{equation*}
\tilde{\mathbf{u}}^{(k)}(0)=\mathrm{TransCNN}_k(\mathbf{z}_k).
\end{equation*}
For prediction at any time point $t \in [0,\ \tilde{r}_t]$, TANTE computes $\tilde{\mathbf{u}}(t)\in\mathbb{R}^{H\times W\times D}$ by summing the Taylor series with the estimated derivatives:
\begin{equation*}
\tilde{\mathbf{u}}(t)=\mathbf{u}(0) + \sum_{k=1}^{n} {\tilde{\mathbf{u}}^{(k)}(0)\cdot t^{k}}/{k!}\ .
\end{equation*}
This neural Taylor expansion enables temporally continuous predictions within the confidence interval without rerunning the network. For predictions beyond $\tilde{r}_t$, TANTE operates autoregressively, concatenating the input with previously predicted states and rerunning the model with the new sequence. All the relevant terms above have been provided in Appendix~\ref{Nomenclature}, and further discussions of the model design are provided in Appendix~\ref{theoretical-practical}.

\section{Results}
\label{Experiments}

We evaluate our proposed TANTE models on The-Well dataset \cite{ohana2024thewell}, comparing their performance against a broad set of state-of-the-art operator learning baselines. Our analysis highlights (a) TANTE’s superior predictive accuracy; (b) parameter efficiency, scalability, and effects of different expansion orders across small, medium, and large configurations; (c) inference efficiency via adaptive rollout; and (d) robustness to key hyperparameters through ablation studies. Importantly, we analyze the confidence radius of convergence ($\tilde{r}_t$) estimated by TANTE, revealing how it systematically adapts to both physical parameters and temporal dynamics, demonstrating TANTE's ability to intelligently adjust its temporal resolution based on the underlying complexity of system evolution.

\paragraph{Benchmarks.}
We adopt four challenging benchmarks, each presenting distinct multiscale and nonlinear dynamics. The full details on the data splits, underlying equations, dataset generation, and problem setup for each case are provided in Appendix~\ref{app-benchmarks}.
\begin{itemize}[leftmargin=*]
    \item \textbf{Turbulent Radiative Layer (TR)}~\cite{fielding2020multiphase}:  Turbulent mixing between cold dense gas clumps and hot ambient gas generates rapidly cooling intermediate-temperature regions, where the competition between radiative energy loss and turbulent velocity fields nonlinearly regulates cold phase growth or dissolution.
    \item \textbf{Active Matter (AM)} \cite{maddu2024learning}: Active matter systems, composed of energy-transforming agents that generate orientation-dependent viscosity and transmit forces, exhibit complex nonlinear spatiotemporal dynamics in viscous fluids.
    \item \textbf{Viscoelastic Fluids (VF)} \cite{beneitez2024multistability}:  Viscoelastic FENE-P fluid flow in wall-bounded geometries, resolving coupled Navier-Stokes and nonlinear conformation tensor dynamics to study multiscale elasto-inertial phenomena.
    \item \textbf{Rayleigh-Bénard Convection (RB)} \cite{burns2020dedalus}: A buoyancy-driven turbulent flow arising from thermally induced density gradients in fluid layers bounded by contrasting thermal boundary conditions, exhibits nonlinear multiscale transport phenomena critical to geophysical, astrophysical, and engineered systems.
\end{itemize}

\paragraph{TANTE model setup.}
In the main experiments, we evaluate two TANTE variants: TANTE-1 and TANTE-2, with maximum derivative orders $n \in \{1,2\}$ at a small model size. Full architectural specifications are provided in Appendix~\ref{model-details}. To assess the benefits of time adaptivity, we also include a fixed-step variant, TANTE-0, which serves as a fixed-step baseline. TANTE-0 is equivalent to TANTE-1 but only predicts the first-order derivative $\tilde{\mathbf{u}}^{(1)}(0) \in \mathbb{R}^{H \times W \times D}$, without estimating a convergence radius. It produces a single-step prediction as
\begin{equation*}
    \tilde{\mathbf{u}}(1)=\mathbf{u}(0)+\tilde{\mathbf{u}}^{(1)}(0),
\end{equation*}
and does not perform time-adaptive rollout. By comparing TANTE-0 to the adaptive TANTE-1, we directly quantify the gains in accuracy and computational efficiency introduced by our adaptive mechanism.

\paragraph{Baselines.}
We benchmark TANTE against competitive operator learning baselines, including Fourier Neural Operator (FNO)~\cite{li2020fourier},  Tensorized Fourier Neural Operator (TFNO)~\cite{kossaifi2023multigridtensorizedfourierneural}, Adaptive Fourier Neural Operator (AFNO)~\cite{guibas2022adaptivefourierneuraloperators}, U-Shaped Neural Operator (UNO)~\cite{rahman2023uno}, and Denoising Pre-training Operator Transformer (DPOT)~\cite{hao2024dpot}.
We also compare against vision-based models designed for operator learning that achieve strong performance: ConvNeXt U-Net (CNextUNet)~\cite{Liu_2022_CVPR}, Attention U-Net (AttUNet)~\cite{oktay2018attentionunetlearninglook}, Axial Vision Transformer (AViT)~\cite{NEURIPS2024_d7cb9db5}, and Continuous Vision transformer (CViT)~\cite{wang2025cvitcontinuousvisiontransformer}, while CViT has been shown to outperform prior baselines across a wide range of PDE benchmarks. All baseline implementations follow the configurations recommended in their respective papers. The rationale for baseline selection is provided in Appendix~\ref{baseline-choice}.

\subsection{Problem setup} 

We adopt the problem setup from the dataset \emph{The Well}~\cite{ohana2024thewell}. All models receive input data $\mathbf{u}_{\text {in }} \in \mathbb{R}^{T\times H\times W\times D}$ over $T=4$ historical timestamps, and predict an output sequence $\mathbf{\tilde{u}}_{\text {pred }} \in \mathbb{R}^{T'\times H\times W\times D}$ for the next $T'$ time steps. When the model’s default output length is shorter than $T'$, it performs an auto-regressive rollout to generate the remaining time steps. 

\paragraph{Training.}
During training, models perform a four-step prediction and minimize the average Mean Squared Error (MSE) between the model predictions $\mathbf{\tilde{u}}_{\text {pred }}$ and the corresponding targets $\mathbf{u}_{\text {true }}$ at the predicted time steps, averaged over all spatial coordinates:
\begin{equation*}
    \mathrm{MSE} = \frac{1}{D}\sum_{d=1}^{D}\left| \vphantom{\sum} \mathbf{\tilde{u}}_{\text {pred }}^d-\mathbf{u}_{\text {true }}^d \right|_2^2,
\end{equation*}

where $\mathbf{u}_{\text {true }}^d$ denotes the $d$-th channel of all time steps in the target sequence $\mathbf{u}_{\text {true }}$, and $\mathbf{\tilde{u}}_{\text {pred }}^d$ denotes the corresponding model prediction in $\mathbf{\tilde{u}}_{\text {pred }}$. For TANTE, the additional $\mathcal{L}_{r}$ for regularizing convergence radius is included. Details about the specific training implementation for each model are provided in Appendix~\ref{model-details}.

We use a unified training recipe for all models on the four benchmarks. We train these models for $100,000$ iterations using the AdamW optimizer \cite{loshchilov2019decoupledweightdecayregularization} with a weight decay of $10^{-5}$. The learning rate schedule consists of a linear warm-up over $5{,}000$ iterations from zero to $5\times10^{-5}$, followed by an exponential decay at a rate of $0.9$ for every $5{,}000$ steps. Batch sizes range from $4$ to $32$, depending on problem size and GPU memory constraints. 
All experiments are conducted on a single NVIDIA H100 GPU for 1--9 hours. Detailed computational costs are reported in Appendix~\ref{computational-cost}.

\paragraph{Evaluation.}
After training, we obtain the predicted trajectory $\mathbf{\tilde{u}}_{\text {pred }}$ by performing an auto-regressive rollout with length $T'=4$ or $T'=8$ on the test datasets. All evaluation workloads are executed on the NVIDIA H100 GPU used for training. Besides the MSE metric, we evaluate model accuracy using the commonly-used relative $L^2$ error:
\begin{equation*}
    \mathrm{Rel.}\ L^2\ \mathrm{Error} = \frac{1}{D}\sum_{d=1}^{D}\frac{\|\mathbf{\tilde{u}}_{\text {pred }}^d-\mathbf{u}_{\text {true }}^d\|_2}{\|\mathbf{u}_{\text {true }}^d\|_2},
\end{equation*}
where the $L^2$ norm is computed over the rollout prediction at all grid points, averaged over each variable of interest.

\begin{table}[htb]
\centering
\caption{Relative $L^2$ Error (↓ lower is better) of rollouts with standard deviations on 4 time steps on four benchmarks: Turbulent Radiative Layer (TR), Active Matter (AM), Viscoelastic Fluids (VF), and Rayleigh-Bénard Convection (RB). The best, second-best, and third-best results are shown in \textbf{bold} and awarded \Gold, \Silver \ and \Bronze \ separately. We mark our TANTE models with \textcolor{red!68!black}{red font}.}
\vspace{1mm}
\label{rel_4}
\resizebox{\textwidth}{!}{
\begin{tabular}{l c
                c@{\hspace{1pt}}c
                c@{\hspace{1pt}}c
                c@{\hspace{1pt}}c
                c@{\hspace{1pt}}c}
\toprule
\textbf{Model} & \textbf{\# Params} & \textbf{TR} & & \textbf{AM} & & \textbf{VF} & & \textbf{RB} & \\
\midrule
FNO      & 4M   & 0.1751 $\pm$ 0.0108 & & 0.3145 $\pm$ 0.0332 & & 0.1955 $\pm$ 0.0135 & & 0.1817 $\pm$ 0.0163 & \\
TFNO     & 4M   & 0.1755 $\pm$ 0.0109 & & 0.3133 $\pm$ 0.0330 & & 0.1970 $\pm$ 0.0135 & & 0.1814 $\pm$ 0.0163 & \\
AFNO     & 5M   & 0.1090 $\pm$ 0.0057 & & 0.1179 $\pm$ 0.0046 & & 0.2840 $\pm$ 0.0121 & & 0.0799 $\pm$ 0.0045 & \\
UNO      & 30M  & 0.1282 $\pm$ 0.0087 & & 0.1128 $\pm$ 0.0052 & & 0.2959 $\pm$ 0.0007 & & 0.4243 $\pm$ 0.0245 & \\
\noalign{\vskip 2pt}
CNextUNet& 4M   & 0.1049 $\pm$ 0.0048 & & 0.1338 $\pm$ 0.0076 & & 0.1723 $\pm$ 0.0123 & & 0.1016 $\pm$ 0.0056 & \\
AttUNet  & 35M  & 0.1056 $\pm$ 0.0070 & & 0.1120 $\pm$ 0.0082 & & 0.0818 $\pm$ 0.0023 & & 0.0745 $\pm$ 0.0043 & \\
\noalign{\vskip 2pt}
DPOT-S   & 32M  & 0.1274 $\pm$ 0.0065 & & 0.2243 $\pm$ 0.0081 & & 0.2670 $\pm$ 0.0121 & & 0.0742 $\pm$ 0.0033 & \\
DPOT-M   & 112M & 0.1119 $\pm$ 0.0060 & & 0.2071 $\pm$ 0.0051 & & 0.2388 $\pm$ 0.0122 & & 0.0836 $\pm$ 0.0054 & \\
DPOT-L   & 540M & 0.1175 $\pm$ 0.0056 & & 0.2431 $\pm$ 0.0136 & & 0.1970 $\pm$ 0.0084 & & 0.0975 $\pm$ 0.0069 & \\
\noalign{\vskip 2pt}
AViT     & 29M  & 0.1264 $\pm$ 0.0070 & & 0.1539 $\pm$ 0.0088 & & 0.0443 $\pm$ 0.0016 & & 0.0933 $\pm$ 0.0062 & \\
CViT     & 31M  & \textbf{0.0882 $\pm$ 0.0061} & \Bronze & \textbf{0.0375 $\pm$ 0.0011} & \Bronze & 0.2315 $\pm$ 0.0140 & & 0.2891 $\pm$ 0.0266 & \\
\hline
\noalign{\vskip 3pt}
\textcolor{red!68!black}{TANTE-0}   & 4.41M   & 0.1088 $\pm$ 0.0076 & & 0.0437 $\pm$ 0.0013 & & \textbf{0.0310 $\pm$ 0.0052} & \Bronze & \textbf{0.0337 $\pm$ 0.0015} & \Bronze \\
\textcolor{red!68!black}{TANTE-1}   & 4.52M   & \textbf{0.0790 $\pm$ 0.0046} & \Silver & \textbf{0.0286 $\pm$ 0.0009} & \Gold & \textbf{0.0270 $\pm$ 0.0043} & \Silver & \textbf{0.0285 $\pm$ 0.0261} & \Gold \\
\textcolor{red!68!black}{TANTE-2}   & 4.81M   & \textbf{0.0660 $\pm$ 0.0036} & \Gold & \textbf{0.0297 $\pm$ 0.0009} & \Silver & \textbf{0.0263 $\pm$ 0.0044} & \Gold & \textbf{0.0335 $\pm$ 0.0015} & \Silver \\
\bottomrule
\end{tabular}
}
\end{table}

\begin{table}[htb]
\centering
\caption{Relative $L^2$ error (↓ lower is better) of rollouts with standard deviations on 8 time steps on four benchmarks: Turbulent Radiative Layer (TR), Active Matter (AM), Viscoelastic Fluids (VF), and Rayleigh-Bénard Convection (RB). The best, second-best, and third-best results are shown in \textbf{bold} and awarded \Gold, \Silver \ and \Bronze \ separately. We mark our TANTE models with \textcolor{red!68!black}{red font}.}
\vspace{1mm}
\label{rel_8}
\resizebox{\textwidth}{!}{
\begin{tabular}{l c
                c@{\hspace{1pt}}c
                c@{\hspace{1pt}}c
                c@{\hspace{1pt}}c
                c@{\hspace{1pt}}c}
\toprule
\textbf{Model} & \textbf{\# Params} & \textbf{TR} & & \textbf{AM} & & \textbf{VF} & & \textbf{RB} & \\
\midrule
FNO      & 4M   & 0.2044 $\pm$ 0.0132 & & 0.5257 $\pm$ 0.0465 & & 0.1903 $\pm$ 0.0127 & & 0.2590 $\pm$ 0.0263 & \\
TFNO     & 4M   & 0.2055 $\pm$ 0.0134 & & 0.5231 $\pm$ 0.0458 & & 0.1922 $\pm$ 0.0126 & & 0.2588 $\pm$ 0.0316 & \\
AFNO     & 5M   & 0.1460 $\pm$ 0.0094 & & 0.2687 $\pm$ 0.0184 & & 0.2835 $\pm$ 0.0105 & & 0.1234 $\pm$ 0.0082 & \\
UNO      & 30M  & 0.1556 $\pm$ 0.0122 & & 0.2683 $\pm$ 0.0246 & & 0.3397 $\pm$ 0.0014 & & 0.6825 $\pm$ 0.0950 & \\
\noalign{\vskip 2pt}
CNextUNet& 4M   & 0.1328 $\pm$ 0.0068 & & 0.3223 $\pm$ 0.0258 & & 0.1727 $\pm$ 0.0125 & & 0.1577 $\pm$ 0.0108 & \\
AttUNet  & 35M  & 0.1458 $\pm$ 0.0107 & & 0.3226 $\pm$ 0.0371 & & 0.1461 $\pm$ 0.0066 & & 0.1386 $\pm$ 0.0112 & \\
\noalign{\vskip 2pt}
DPOT-S   & 32M  & 0.1537 $\pm$ 0.0090 & & 0.4355 $\pm$ 0.0273 & & 0.2674 $\pm$ 0.0091 & & 0.1093 $\pm$ 0.0066 & \\
DPOT-M   & 112M & 0.1376 $\pm$ 0.0082 & & 0.3690 $\pm$ 0.0197 & & 0.2371 $\pm$ 0.0116 & & 0.1426 $\pm$ 0.0134 & \\
DPOT-L   & 540M & 0.1447 $\pm$ 0.0079 & & 0.3830 $\pm$ 0.0237 & & 0.2164 $\pm$ 0.0101 & & 0.1682 $\pm$ 0.0182 & \\
\noalign{\vskip 2pt}
AViT     & 29M  & 0.1677 $\pm$ 0.0117 & & 0.3583 $\pm$ 0.0394 & & 0.0801 $\pm$ 0.0034 & & 0.1490 $\pm$ 0.0138 & \\
CViT     & 31M  & \textbf{0.1174 $\pm$ 0.0097} & \Bronze & 0.1077 $\pm$ 0.0106 & & 0.2143 $\pm$ 0.0114 & & 0.3153 $\pm$ 0.0285 & \\
\hline
\noalign{\vskip 3pt}
\textcolor{red!68!black}{TANTE-0}   & 4.41M   & 0.1346 $\pm$ 0.0101 & & \textbf{0.0956 $\pm$ 0.0061} & \Bronze & \textbf{0.0519 $\pm$ 0.0086} & \Bronze & \textbf{0.0707 $\pm$ 0.0046} & \Bronze \\
\textcolor{red!68!black}{TANTE-1}   & 4.52M   & \textbf{0.1106 $\pm$ 0.0078} & \Silver & \textbf{0.0767 $\pm$ 0.0063} & \Gold & \textbf{0.0457 $\pm$ 0.0075} & \Silver & \textbf{0.0584 $\pm$ 0.0035} & \Gold \\
\textcolor{red!68!black}{TANTE-2}   & 4.81M   & \textbf{0.0988 $\pm$ 0.0073} & \Gold & \textbf{0.0804 $\pm$ 0.0072} & \Silver & \textbf{0.0403 $\pm$ 0.0076} & \Gold & \textbf{0.0598 $\pm$ 0.0044} & \Silver \\
\bottomrule
\end{tabular}
}
\end{table}

\subsection{Main results}
\label{main-results}

Tables~\ref{rel_4} and~\ref{rel_8} present two comprehensive comparisons of TANTE models against competitive baselines when the rollout length $T'=4$ and $T'=8$. Additional results with the MSE metric are provided in Appendix~\ref{additional-results}. As shown in these tables, our proposed method achieves the lowest relative $L^2$ error on all benchmarks, with TANTE-1 consistently outperforming its fixed-step counterpart TANTE-0. This demonstrates the effectiveness of our time-adaptive module in improving prediction accuracy, an advantage that becomes especially pronounced during rollouts where the accumulation of local errors degrades the performance of fixed-step models. Additional rollout trajectory visualizations are shown in Figure~\ref{main_viz_3}, generated by the best-performing TANTE variant from Table~\ref{rel_4}. These results further illustrate TANTE’s superior performance.

\begin{figure}[htbp]
\begin{center}
\centerline{\includegraphics[width=\textwidth]{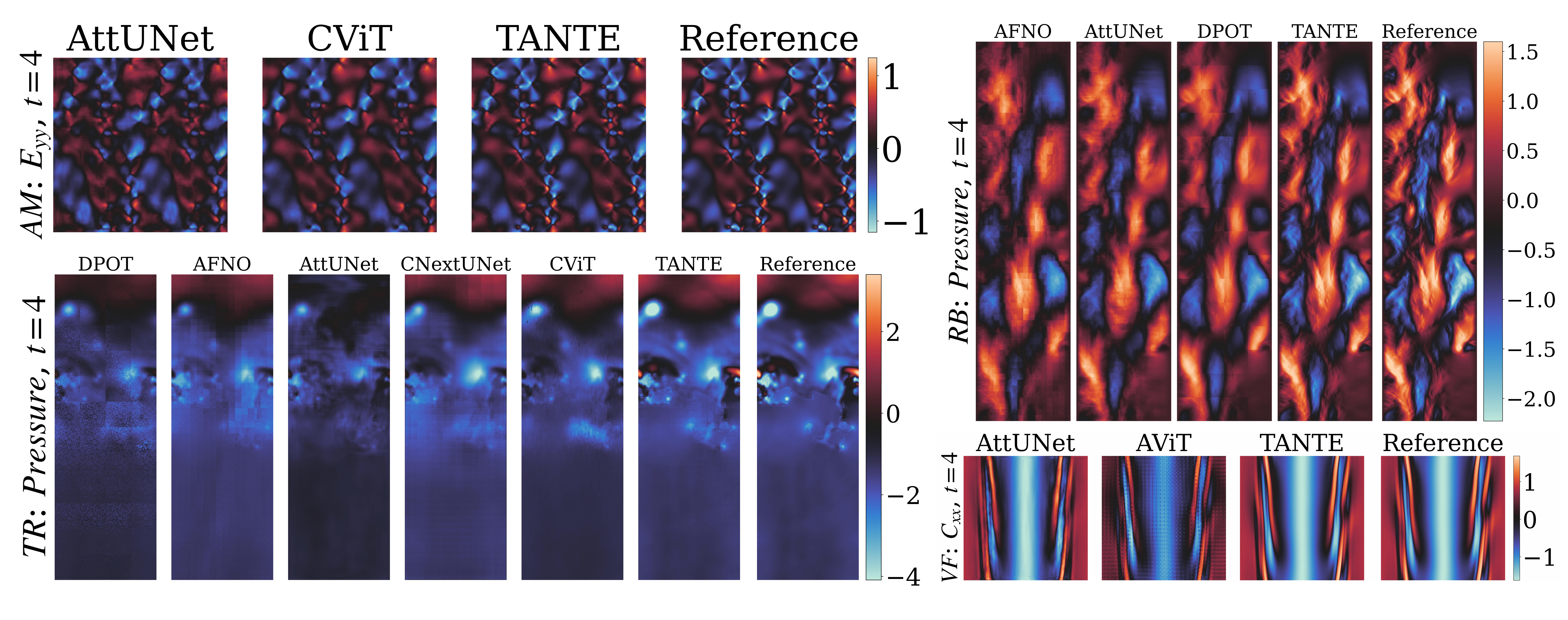}}
\caption{Predictions of the target field at $t=4$ on the four benchmarks. 
For each dataset, we show one representative sample comparing our approach with the best performance against several competitive baselines with top accuracy.
}
\label{baseline_viz}
\end{center}
\vspace{-6mm}
\end{figure}

We also visualize one representative sample from each benchmark in Figure~\ref{baseline_viz}, comparing TANTE's predictions with those of the top baselines.
We observe that TANTE introduces the minimal checkboard artifacts among patch-based models, yielding outputs that are nearly continuous. Furthermore, while the CViT model uses continuous representations, TANTE preserves finer details and exhibits fewer additional artifacts. These visualizations further highlight TANTE’s capacity to model complex physical fields while maintaining fine details during prediction.

It is worth noting that the baseline Axial Vision Transformer (AViT)~\cite{NEURIPS2024_d7cb9db5} shares a key architectural idea with TANTE. Unlike conventional Vision-Transformer style models, AViT and TANTE do not flatten the encoded input $\mathbf{u}_{st}\in\mathbb{R}^{T\times H'\times W'\times C}$ into a single sequence of length $T H' W'$, or into $T$ sequences of length $H' W'$ to compute attention. Instead, they compute attention separately along each axis—time, height, and width. This shifts the attention complexity from $\mathcal{O}\big((T H' W')^{2}\big)$ for fully flattened attention, or $\mathcal{O}\big((H' W')^{2}\big)$ for attention on each frame, to $\mathcal{O}\big(T^{2}+H'^{2}+W'^{2}\big)$. Because the $T$ in our experiments is much smaller than $H'$ and $W'$, this yields a substantial reduction in computational cost. However, a key distinction exists between the Transformer Processor of TANTE and AViT. TANTE stacks independent Transformer blocks, each of which sequentially processes a different axis. Whereas in AViT, the spatial $K$, $Q$, and $V$ projections are shared between the height and width axes~\cite{NEURIPS2024_d7cb9db5}. This weight sharing limits AViT’s capacity to model spatially anisotropic systems. As shown in Tables~\ref{rel_4} and~\ref{rel_8}, AViT underperforms TANTE-0 with the same training recipe, despite having approximately $6\times$ the number of parameters. This comparison underscores the advantage of TANTE's Transformer Processor design. 

\subsection{Error accumulation} 
\label{Error-accumulation}

\begin{figure}[htbp]
\begin{center}
\centerline{\includegraphics[width=\textwidth]{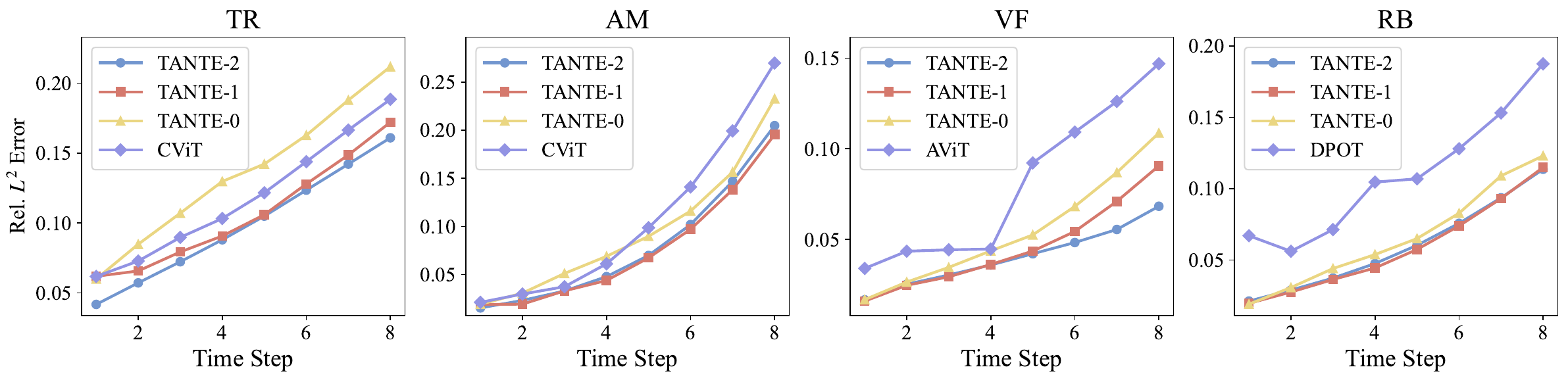}}
\caption{Relative $L^2$ error at each of eight rollout time points for four PDE benchmarks (\textit{TR}, \textit{AM}, \textit{VF}, and \textit{RB}). TANTE-2 (blue) and TANTE-1 (red) show the lowest average error across all time steps and the minimum cumulative error compared to the best baseline methods.}
\label{accumulate}
\end{center}
\vspace{-6mm}
\end{figure}

Error accumulation is a central challenge in operator learning when performing auto-regressive rollout. The models iteratively feed their predictions back as input for future steps. Consequently, errors compound over time—early inaccuracies propagate and amplify, degrading long-horizon forecasts. Therefore, understanding this error propagation mechanism is crucial for evaluating model capabilities and explains why time-adaptive approaches outperform fixed-step methods. To illustrate error accumulation behavior, we compare the temporal error trajectories of TANTE-0, TANTE-1, and TANTE-2 with the best baseline model on each benchmark (Tables~\ref{rel_4} and~\ref{rel_8}). Specifically, we compute the relative $L^2$ relative error for each rollout time step from $t=1$ to $t=8$, averaging the errors independently for each time step (Figure~\ref{accumulate}).

Figure~\ref{accumulate} also demonstrates several key observations. First, there is no significant difference in the error of the first time step between TANTE-0 and TANTE-1, which can be attributed to their identical model architecture. However, the error progression of TANTE-1 remains lower than that of TANTE-0 in subsequent time steps. This highlights the advantages of the time-adaptive approach: by reducing prediction steps during periods of gradual change, TANTE-1 mitigates the systematic errors that would accumulate with each prediction step, stabilizing the error and maintaining a minimal cumulative error over time. Furthermore, TANTE-2 exhibits lower initial error in the \textit{TR} benchmark and demonstrates stronger stability across time steps in the \textit{VF} benchmark, suggesting that second-order expansion provides superior modeling capability compared to first-order expansion in these problems.

\subsection{Scalability}
\label{scalability}

Tables~\ref{rel_4} and~\ref{rel_8} show the superior performances of TANTE-1 and TANTE-2 at a small parameter size. Additionally, TANTE-2 outperforms TANTE-1 across two benchmarks but has slightly higher errors than TANTE-1 on the other two benchmarks. Therefore, we conduct an experiment to evaluate the scalability of TANTE in terms of parameter sizes and expansion orders.

\begin{figure}[htbp]
\begin{center}
\centerline{\includegraphics[width=\textwidth]{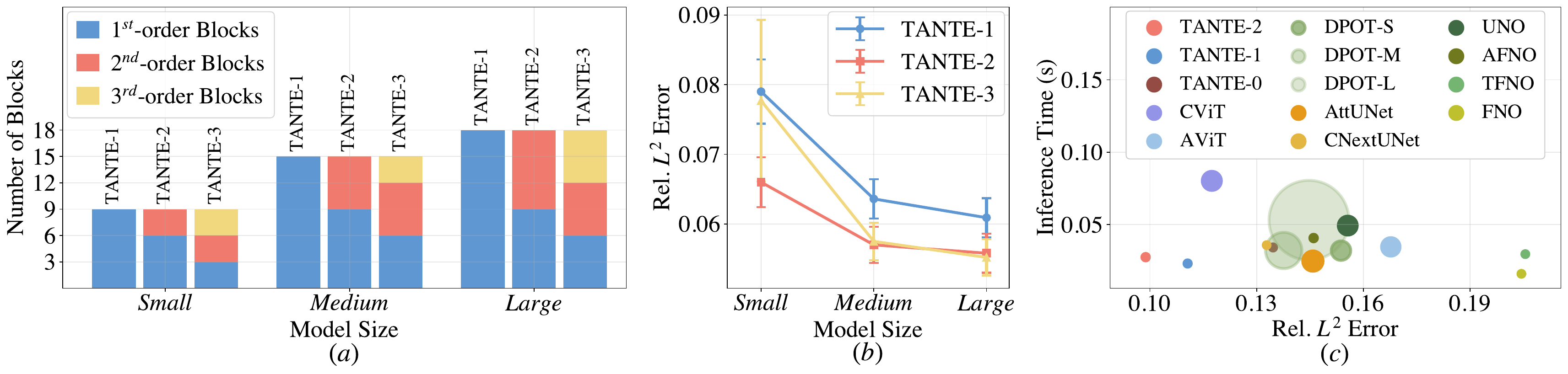}}
\caption{
\textit{(a) Transformer block allocations for TANTE variants at three parameter sizes.} The Transformer blocks in TANTE are divided into $n$ groups to approximate $n$ orders of derivatives, ensuring comparable model sizes at each level of parameter count. Different colors represent blocks used for estimating derivatives at different orders.
\textit{(b) Test errors and standard deviations of TANTE variants at three model sizes on the TR benchmark.} The prediction accuracy positively correlates with the parameter count for each TANTE variant. Although TANTE-3 shows higher error than TANTE-2 at small and medium sizes, it exhibits better scalability and achieves the lowest error at the large parameter size.
(c) \textit{Average inference time at eight-step rollout}. TANTE-1 and TANTE-2 attain the lowest error (highest accuracy) while remaining as fast as other baselines. Additionally, TANTE-1 and TANTE-2 are faster than TANTE-0.
}
\label{scale}
\end{center}
\vspace{-4mm}
\end{figure}

First, we construct TANTE models in three configurations—small, medium, and large—as summarized in Table~\ref{variant_config}. Detailed parameter sizes are reported in Table~\ref{param_size}. Next, at each parameter size, we design three TANTE variants. As shown in Figure~\ref{pipeline}, the Transformer blocks in TANTE's Transformer Processor are partitioned into $n$ groups to form a TANTE model with a maximum expansion order of $n$, denoted as $1^{st}$-, $2^{nd}$-, and $3^{rd}$-order blocks. To ensure that TANTE variants have comparable model sizes, the total number of blocks is fixed at each size level. Figure~\ref{scale}a illustrates the composition of each TANTE variant at the three size levels.

\begin{table}[htbp]
\centering
\caption{TANTE configurations at small, medium, and large sizes. The number of parameters is averaged across TANTE-0, TANTE-1, and TANTE-2.
}
\label{variant_config}
\begin{tabular}{c c c c c c}
\toprule
\textbf{Size} & \textbf{Embed dim.} & \textbf{MLP dim.} & \textbf{\# Heads} & \textbf{\# Blocks} & \textbf{\# Params} \\
\midrule
Small & 256 & 256 & 8  & 9  & 5M  \\
Medium & 384 & 768 & 12 & 15 & 19M \\
Large & 512 & 1024 & 16 & 18 & 40M \\
\bottomrule
\end{tabular}
\vspace{0mm}
\end{table}

As the maximum expansion order increases at a given parameter size, fewer blocks remain for estimating lower orders, which may reduce the modeling capacity of the lower-order derivatives. If this reduction outweighs the gain from higher-order estimation, the error increases—as seen in TANTE-2 performing worse than TANTE-1 on two benchmarks. However, as the total number of blocks scales up, higher-order TANTE variants regain sufficient blocks for modeling lower-order derivatives, which demonstrates stronger scalability with parameter size. Figure~\ref{scale}b supports this explanation. At the small level of parameter sizes, TANTE-3 is outperformed by TANTE-2 and exhibits error levels similar to TANTE-1, with even larger standard deviations. 
As the parameter size increases to the medium level, the error of TANTE-3 decreases, matching that of TANTE-2 while significantly outperforming TANTE-1.
At large parameter sizes, TANTE-3 surpasses TANTE-2, demonstrating the strongest scalability.

Overall, Figure~\ref{scale} illustrates both the scalability of TANTE variants with respect to parameter sizes and the maximum expansion order, highlighting the advantage of the Taylor-expansion-based approach in decomposing the learning of the target state into manageable components.

\subsection{Inference time} 
\label{Inference-time}
Inference speed is another key consideration in operator learning, especially for rollout tasks. Our proposed time-adaptive approach leverages the estimated local convergence radius ($\tilde{r}_t$) to reduce the number of prediction steps required during rollout when $\tilde{r}_t > 1$, potentially enhancing inference speed without sacrificing accuracy.

We compared the average rollout inference time of TANTE and other baseline models on the \textit{TR} benchmarks with rollout length $T' = 8$. As shown in Figure~\ref{scale}c, TANTE-1 and TANTE-2 achieve the lowest relative $L^2$ errors while maintaining inference speeds comparable to or faster than other baselines. More importantly, both TANTE-1 and TANTE-2 are consistently faster than TANTE-0, despite having nearly identical architecture and parameter size, showing that the time-adaptive module results in fewer steps and thus a better inference efficiency.

We also observe a slight increase in inference time from TANTE-1 to TANTE-2, caused by the additional operations for estimating higher-order derivatives. However, this increase is not significant, alleviating concerns about the robustness of TANTE's inference speed across different variants.

\subsection{Ablation experiments}
\label{Ablation}

\begin{figure}[htbp]
\begin{center}
\centerline{\includegraphics[width=1.0\columnwidth]{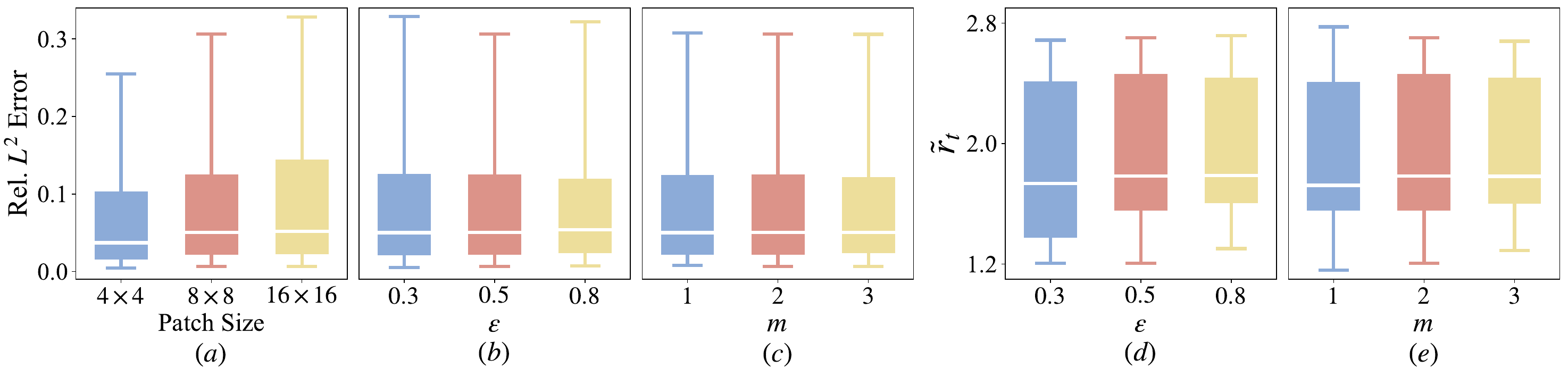}}
\caption{
    \textit{Ablation studies for TANTE on the Turbulent Radiative Layer (TR) benchmark.} Box plots of test errors distributions for (a) different patch sizes, (b) different $\varepsilon$ in $\mathcal{L}_{r}$, and (c) different power $m$ in $\mathcal{L}_{r}$. Distributions of $\tilde{r}_t$ in evaluation for (d) different $\varepsilon$ in $\mathcal{L}_{r}$ and (e) different power $m$ in $\mathcal{L}_{r}$. Results obtained using TANTE-1 with $8\times8$ patch-size, $\varepsilon=0.5$ and $m=2$, varying each hyper-parameter of interest while keeping others fixed.}
\label{ablation}
\end{center}
\vspace{-5mm}
\end{figure}

We perform ablation studies on key hyperparameters of TANTE using the Turbulent Radiative Layer (\textit{TR}) benchmark (Figure~\ref{ablation}). We begin by evaluating the effect of patch size on performance. As shown in Figure~\ref{ablation}a, smaller patch sizes generally lead to better accuracy, but at the cost of increased computational overhead.

Next, we investigate the impact of the regularization term $\mathcal{L}_{r}$, introduced in Section~\ref{Method} to penalize small values of $\tilde{r}_t$. Specifically, we ablate its two hyperparameters: the penalty threshold $\varepsilon$ and the exponent $m$. Figures~\ref{ablation}b--e show that varying $\varepsilon$ and $m$ has little effect on predictive accuracy or distribution of $\tilde{r}_t$, suggesting that the accuracy of TANTE is robust to these choices. 

Intuitively, increasing the penalty threshold $\varepsilon$ should shift the learned $\tilde{r}_t$ upward. Consistent with this expectation, Figure~\ref{ablation}d shows a modest increase in the minimum and lower quartile of $\tilde{r}_t$ as $\varepsilon$ grows, while the median and other summary statistics remain essentially unchanged. Likewise, Figure~\ref{ablation}e indicates that larger exponents $m$ produce a more concentrated distribution, while leaving the median mostly unaffected.

Overall, these ablation studies validate our default hyperparameter settings and highlight practical design considerations for future time-adaptive operator learning models.

\subsection{Analysis of adaptive radius of convergence}

TANTE estimates the confidence radius of convergence $\tilde{r}_t$, which adapts dynamically across rollout steps. In this section, we analyze how these variations relate to the underlying system properties of the temporal evolution.

\subsubsection{Adaptivity across system parameters}
\label{cross-param}
To begin with, we examine how $\tilde{r}_t$ varies across different subsets of the same dataset, each defined by a distinct physical parameter.  
These PDE parameters are closely correlated with the system's complexity. We focus on two key parameters: the dimensionless active dipole strength $\alpha$ in the Active Matter (\textit{AM}) benchmark, and the Rayleigh number $Ra$ in the Rayleigh-Bénard Convection (\textit{RB}) benchmark. We explore the relationship between these parameters and system complexity, as well as the correlation between the estimated $\tilde{r}_t$ and the physical parameters.

\paragraph{Active dipole strength.}
The parameter $\alpha$ represents the dimensionless active dipole strength in the Active Matter system. A negative $\alpha$ means repulsive alignment interactions between particles. The active dipole strength influences the governing PDEs through its role in the stress tensor $\mathbf{\Sigma}$, thereby affecting particle alignment, viscosity, and the overall flow dynamics in the system. As $\alpha$ becomes more negative, the repulsive alignment interactions intensify, reducing the collective motion of particles. This makes the system rely more on random diffusion than coordinated active behavior, slowing down the evolution. Details are provided in Appendix~\ref{am_details}.

\paragraph{Rayleigh number.} 
The Rayleigh number, $Ra$, influences the PDEs of the \textit{RB} system through its impact on thermal diffusivity $\kappa$ and viscosity $\nu$:
\begin{equation*}
    \kappa = (Ra\cdot Pr)^{-1/2},\quad
    \nu = (Ra/ Pr)^{-1/2},
\end{equation*}
where $Pr$ is the fixed Prandtl number. An increase in the Rayleigh number $Ra$ results in lower values for both thermal diffusivity $\kappa$ and viscosity $\nu$. A lower $\kappa$ makes heat become more trapped in localized warm regions, allowing the formation of significant temperature gradients. This strengthens buoyancy forces, promoting vigorous convection and accelerating the system's evolution. Additionally, a reduction in $\nu$ makes the fluid less resistant to flow, allowing buoyant forces to overcome viscous resistance more easily, thereby promoting rapid and potentially turbulent convective motion. In summary, higher Rayleigh numbers accelerate convection and increase the rate of system evolution.

\paragraph{Spectral entropy.}
To support our assertion regarding the correlation between system parameters and their complexity in temporal evolution, we computed the average Spectral entropy (SE) for the Active Matter and Rayleigh-Bénard Convection datasets under each of the parameters. Spectral entropy is a measure derived from information theory, quantifying the complexity or unpredictability of a time series. Mathematically, the calculation of Spectral entropy proceeds as follows. First, we perform the Fourier transform along the time axis for the discrete input tensor $\mathbf{u}\in \mathbb{R}^{T\times H\times W\times D}$, obtaining the frequency-domain representation $\mathbf{\hat{u}} = \mathcal{F}\ \mathbf{u}$. We take the squared magnitude of $\mathbf{\hat{u}}\in\mathbb{C}^{T\times H\times W\times D}$ to obtain the Power Spectrum $\mathbf{p} = |\mathbf{\hat{u}}|^2\in\mathbb{R}^{T\times H\times W\times D}$. Next, we normalize $\mathbf{p}$ along the time axis to form a probability distribution $\mathbf{P}$, and compute its Shannon entropy $SE$:
\begin{equation*}
    \mathbf{P} = \frac{\mathbf{p}}{\sum_{t=1}^{T} \mathbf{p}_t}\ ;\quad 
    SE=-\sum_{t=1}^{T}\mathbf{P}_t\cdot \log_2 \mathbf{P}_t/\log_2 T.
\end{equation*}
The final spectral entropy $SE$ is averaged over all the dimensions to quantifies the overall flatness of the power spectrum. A higher $SE$ indicates a more uniform distribution of power across frequencies, corresponding to greater complexity or randomness in the time series.

The results in Table~\ref{cpl_se} confirm our statements. The complexity measured by spectral entropy values of benchmarks aligns with our assertions: larger $|\alpha|$ correspond to more slowly changing dynamics, while higher Rayleigh numbers are associated with more rapidly changing flows.

\begin{table}[htbp]
\centering
\caption{Spectral entropy ($SE$) values for subsets of the Active Matter and Rayleigh-Bénard Convection datasets.}
\label{spectral_entropy}
\setlength{\tabcolsep}{5pt}
\begin{minipage}[t]{0.48\linewidth}
\centering
\textbf{AM - Active dipole strength}\\[2pt]
\begin{tabular}{cccccc}
\toprule
$\alpha$ & $-1.0$ & $-2.0$ & $-3.0$ & $-4.0$ & $-5.0$ \\
\midrule
$SE$ & 0.7376 & 0.6922 & 0.6744 & 0.6674 & 0.6641 \\
\bottomrule
\end{tabular}
\end{minipage}
\hfill
\begin{minipage}[t]{0.48\linewidth}
\centering
\textbf{RB - Rayleigh number}\\[2pt]
\begin{tabular}{cccccc}
\toprule
$Ra$ & $10^6$  & $10^7$  & $10^8$  & $10^9$  & $10^{10}$ \\ \midrule
$SE$ & 0.7003 & 0.7169 & 0.7382 & 0.7494 & 0.7585 \\ \bottomrule
  \end{tabular}
\end{minipage}
\label{cpl_se}
\vspace{-3mm}
\end{table}

\paragraph{$\tilde{r}_t$ vs. system parameters.}

\begin{figure}[htbp]
\begin{center}
\centerline{\includegraphics[width=1.0\columnwidth]{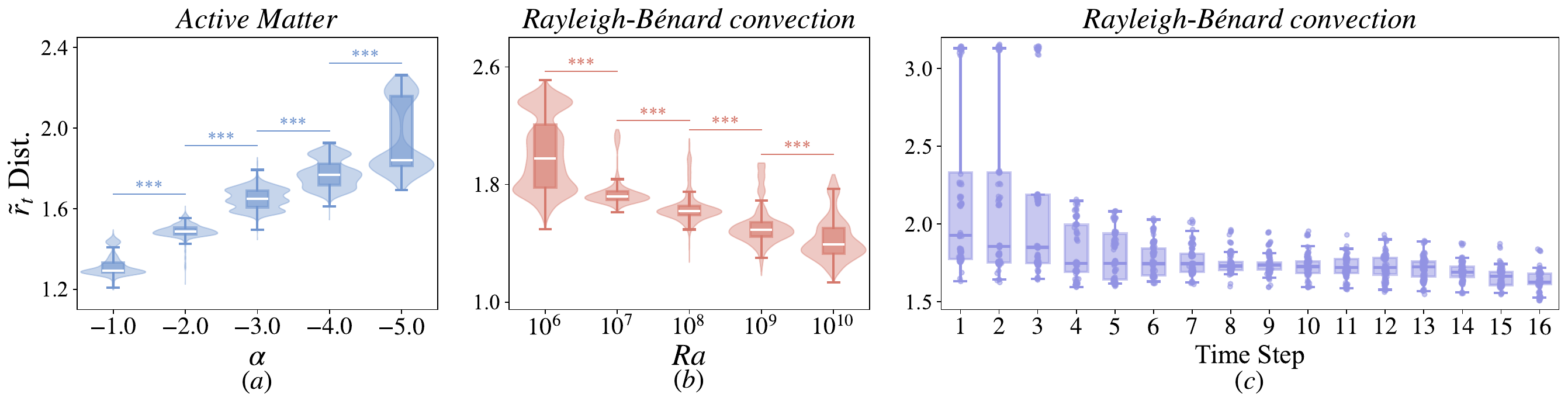}}
\caption{
\textit{Distributions of TANTE's estimated $\tilde{r}_t$ to illustrate how TANTE adaptively adjusts its temporal resolution throughout the simulation.} 
(a--b) Distributions of TANTE’s estimated $\tilde{r}_t$ across different parameters in the \textit{AM} and \textit{RB} benchmarks. The $p$‐values from a Mann‐Whitney \textit{U} test (marked as ***, **, *) indicate significant differences in $\tilde{r}_t$ between adjacent parameter settings. As the parameters change, TANTE’s estimated $\tilde{r}_t$ shifts accordingly, demonstrating the model’s ability to adapt its temporal resolution to the underlying dynamical complexity. 
(c) TANTE’s estimated $\tilde{r}_t$ distributions steps during the initial phase of the \textit{RB} benchmark. As the predictive horizon extends and the system’s evolution becomes more complex, TANTE gradually reduces the step size. 
Results are obtained using TANTE-1.
}
\label{adaptivity}
\end{center}
\vspace{-6mm}
\end{figure}

As shown in Figure~\ref{adaptivity}a--b, the distribution of $\tilde{r}_t$ values differs significantly between parameter regimes across the \textit{AM} and \textit{RB} benchmarks. Statistical tests (Mann–Whitney \textit{U}, Appendix~\ref{u-test}; annotated as ***, **, *) confirm that TANTE meaningfully distinguishes between subsets based on system dynamics.
As the complexity of the temporal evolution increases, TANTE predicts smaller $\tilde{r}_t$, indicating more conservative step sizes. Conversely, in simpler regimes, larger $\tilde{r}_t$ reflects more confident, longer-step predictions. These patterns suggest that TANTE implicitly learns to adjust its temporal resolution based on the complexity of the governing dynamics.

\subsubsection{Temporal adaptivity within trajectories}
\label{cross-region}
We further analyze TANTE's adaptive behavior across different temporal regions within the same dynamical trajectories. We select the initial stages of a representative benchmark—Rayleigh-Bénard Convection (\textit{RB})—performing rollouts of length $T'=16$. In the system, dynamics evolve from stable initial conditions to growing perturbations, with the complexity of temporal evolution gradually increasing (a sample is visualized in Figure~\ref{main_viz_3} Top Left).

Figure~\ref{adaptivity}c shows the distribution of $\tilde{r}_t$ values over each time step during rollout, averaged across different trajectories. 
TANTE gradually reduces its estimated $\tilde{r}_t$ as prediction proceeds in this benchmark. As evolution complexity increases, TANTE becomes more conservative to maintain accuracy.
This demonstrates that TANTE not only estimates appropriate convergence radii for different systems but also dynamically adjusts step sizes during rollout. Such adaptivity enables it to balance accuracy and efficiency, particularly in long-horizon rollouts where system behavior evolves significantly.

\section{Discussion}

We introduce neural Taylor expansion, a novel framework for operator learning that leverages Taylor series to perform time-adaptive rollout. Building on this foundation, we propose the Time-Adaptive Transformer with Neural Taylor Expansion (TANTE), which achieves state-of-the-art predictive accuracy and parameter efficiency across four challenging PDE benchmarks, and consistently outperforms a wide range of fixed-step baselines. Additionally, we demonstrate its scalability with respect to both parameter size and the maximum order of Taylor expansion. 
Furthermore, TANTE dynamically adapts its temporal resolution to the temporal complexity of the underlying dynamics, adjusting both across distinct physical regimes and within individual trajectories. Collectively, our results establish Time-Adaptive Neural Taylor Expansion as a powerful and generalizable approach for learning time-adaptive surrogate models of complex dynamical systems, with TANTE offering a practical, scalable implementation suitable for real-world applications.

While our work opens exciting avenues for advancing operator learning and scientific machine learning, several limitations deserve further investigation. 
First, Figures~\ref{main_viz_3} and~\ref{baseline_viz} show checkboard artifacts in TANTE's predictions, which arise from the patch-based encoder and decoder. It could be resolved by either incorporating Fourier layers in FNO~\cite{li2020fourier} or integrating the continuous representations introduced by the CViT~\cite{wang2025cvitcontinuousvisiontransformer}.
Second, current experiments focus on systems with regular geometries and uniform grids, leaving performance on complex geometries (e.g., fractured porous media, turbulent multiphase flows) unexplored. We will extend our work to handle such geometries by employing non-uniform fast Fourier transform (NUFFT) with Fourier layers, or some neural surrogates developed for irregular domains.
Third, the current implementation is designed as a specialized operator learning model rather than a generalizable framework, which limits its direct applicability to PDE systems that require coupled multi-physics modeling. We plan to generalize our model by pretraining one Transformer Processor with specialized encoders and decoders, building a unified multimodal foundation model capable of handling diverse field shapes. 
Additionally, incorporating physics-informed constraints or hybrid symbolic-neural methods could enhance both interpretability and physical consistency. We anticipate these directions will catalyze the development of next-generation, adaptive surrogate models that empower more accurate, efficient, and robust simulations across scientific and engineering domains.

\section*{Acknowledgments}

This work was supported in part by the U.S. National Science Foundation (NSF) under CAREER Award No. 2443528, Grant No.~DMS-2347833, and Grant No.~DMS-2527294, and the U.S. Department of Energy Office of Advanced Scientific Computing Research under Grants No.~DE-SC0025593 and No.~DE-SC0025592.

\appendix
\section{Nomenclature}
\label{Nomenclature}

Table \ref{nomenclature} summarizes the main symbols and notation used in this work.

\begin{table}[htbp]
\centering
\renewcommand{\arraystretch}{1.2}
\caption{Summary of the main symbols and notations used in this work.}
\label{nomenclature}
\begin{tabular}{|>{\centering\arraybackslash}m{0.2\textwidth}|>{\arraybackslash}m{0.7\textwidth}|}
\hline
\rowcolor{gray!30}
\textbf{Notation} & \textbf{Description} \\ \hline
\multicolumn{2}{|>{\columncolor{gray!15}}c|}{\textbf{Operator learning}} \\ \hline
$\mathbf{u}, \mathbf{u}_{\text{in}}$ & Spatiotemporal input tensor \\
$\mathbf{u}_{\text{true}}$, $\mathbf{\tilde{u}}_{\text{pred}}$ & Ground-truth and the predicted output tensor \\
$\mathbf{u}_{\text{true}}^d$, $\mathbf{\tilde{u}}_{\text{pred}}^d$ & Ground truth and the predicted field for the $d$-th channel\\
$\mathbf{S}_i$ & $i$-th spatial frames of the input tensor \\
$\mathbf{T}_{\text{seq}}$ & Sequence of relative timestamps of the input tensor \\
$T$, $T'$ & Input / Output sequence length \\
$H\times W$ & Resolution of spatial discretization: Height $\times$ Width \\
$\mathcal{F \cdot}$ & Fourier Transform \\
\hline
\multicolumn{2}{|>{\columncolor{gray!15}}c|}{\textbf{TANTE}} \\ \hline
$\mathbf{u}_p$ & Patchified inputs \\
$\mathbf{u}_s$ & Patchified inputs with positional embedding \\
$\mathbf{u}_{st}$ & Patchified inputs with positional embedding and time encoding \\
$\gamma(\cdot)$, $\beta(\cdot)$ & Learnable scaling/shifting factor derived from the condition $t$. \\
$\mathbf{z}$ & Representation of estimated temporal derivatives at $t=0$ \\
$\mathbf{z}_k$ & Representation of  estimated $k$-th order temporal derivative at $t=0$ \\
$\tilde{r}_t$ & Estimated local time radius of convergence / Adaptive time step size \\
$[0,\ \tilde{r}_t]$ & Confidence interval \\
$\tilde{\mathbf{u}}^{(k)}(0)$ & Estimated $k$th-order temporal derivative \\
$\mathbf{u}(t)$, $\tilde{\mathbf{u}}(t)$ & Ground truth and the predicted fields for the time step $t$ \\
$\text{PE}_{s}$ & Spatial positional embedding \\ 
$\mathrm{FiLM}(\cdot)$ & Feature-wise Linear Modulation layer\\
$\mathrm{MLP}_k(\cdot)$ & MLP for determining $\tilde{r}_t$ from the representation of $k$-th order derivative\\
$\mathrm{TransCNN}_k(\cdot)$ & Transposed CNN for reconstruction of the $k$-th order derivative \\
$\mathcal{L}_{r}(\cdot)$ & Regularization loss term penalizing overly small $\tilde{r}_t$ values \\
$D$ & The number of physical channels \\
$C$ & Embedding dimension of Transformer Processor \\
$n$ & Maximum order of the Taylor expansion of TANTE \\
$P$ & Patch size of Spatiotemporal Encoder \\
$H'$, $W'$ & Height and width of patchified inputs: $H/P$ and $W/P$ \\
$\varepsilon$, $m$ & Hyper-parameters in $\mathcal{L}_{r}(\cdot)$ \\
\hline
\multicolumn{2}{|>{\columncolor{gray!15}}c|}{\textbf{Dynamical system}} \\ \hline
$\alpha$ &  Active dipole of the Active Matter system \\
$\mathbf{\Sigma}$ & Stress tensor in the Active Matter system\\
$Ra$, $Pr$ & Rayleigh number and Prandtl number of the Rayleigh–Bénard convection system \\
$\kappa$, $\nu$ & Thermal diffusivity and viscosity of the Rayleigh–Bénard convection system \\
$\mathbf{p}$ & Power Spectrum\\
$\mathbf{P}$ & Normalized Power Spectrum\\
$SE$ & Spectral entropy \\
\hline
\end{tabular}
\end{table}

\section{Theoretical and practical explanations}
\label{theoretical-practical}
Here, we present supplementary explanations and relevant background information to facilitate a clearer understanding of our model design and experimental procedures.

\subsection{Transformer-based model as neural operator}
\label{transformer-operator}

The distinguishing feature of neural operators lies in their ability to learn mappings between infinite-dimensional function spaces. A critical property of neural operators is their discretization invariance: model parameters are shared across both temporal and spatial resolutions, allowing the output to be evaluated at any point in the domain, regardless of the discretization used during training. The work by Kovachki et al.~\cite{kovachki2024neuraloperatorlearningmaps} demonstrates that transformers are a special case of neural operators. TANTE is well-aligned with this framework.

Although the inputs and outputs are discretized as tensors in $\mathbb{R}^{H \times W \times D}$ for numerical implementation, the model inherently learns a continuous operator. These tensors discretize functions on grids while maintaining resolution invariance. More significantly, the self-attention mechanism implicitly parameterizes integral kernels. Each attention head computes a kernel $\kappa(x, y)$ over coordinates, which corresponds to the integral operator $Kv(x) = \int \kappa(x, y) v(y) dy$ as shown in neural operator theory. Nonlinear activations and residual connections further mirror the iterative kernel integration process inherent to neural operators. In the Transformer Processor described in Section~\ref{Transformer-Processor}, TANTE stacks three types of transformer blocks with attention mechanisms along the time, height, and width axes, enabling behaviors as integral operators in the spatiotemporal domain. This characteristic enables evaluation at arbitrary resolutions, which is a hallmark of neural operators.

\subsection{Sub-interval time step}
TANTE is straightforward to take sub-interval time steps with step size $t<1$.
Since TANTE applies a neural Taylor expansion with arbitrary $t$ (time) as a query, we can easily query time in the range of $[0, 1]$, similar to Refs.~\cite{zhu2023fourier,jiang2024fourier,lee2024efficient}. Although TANTE is trained and evaluated on datasets with evenly spaced time steps (i.e., unit intervals) in our experiments, we anticipate that the learned dynamics are comprehensive to interpolate intermediate states accurately by using $t<1$. Additionally, TANTE employs the FiLM layer for the time encoding, allowing the inputs on the uneven spaced time steps. Furthermore, as discussed in Section~\ref{transformer-operator}, TANTE trained on discrete-time data is able to predict sub-step evolution with its discretization invariance.

However, a practical challenge is the lack of ground-truth data at sub-interval time steps in typical benchmarks.  
Common datasets provide solutions at uniform time intervals, so it is challenging to have ground-truth temporal intermediate states for direct training or evaluation.  
This circumstance makes it infeasible to train the model explicitly with $t<1$, since we cannot compute a loss at unseen intermediate times. Similarly, it is hard to rigorously validate the accuracy of sub-step predictions without reference data.  

In summary, TANTE's framework is capable of predicting with sub-interval steps for finer temporal resolution based on its temporal continuity. The implementation only requires adjusting the query time $t$ to the sub interval value.  
We have not formally included sub-interval evaluation in our experiments because of the dataset limitations mentioned, but this is a promising direction for practical application.

\section{Experimental details}
\label{experimental_details}

\subsection{Dataset}
\label{app-benchmarks}

We make use of the datasets The Well~\cite{ohana2024thewell}, which consists of 15 TB data of discretized initial conditions on diverse types and parameter sets. We compare TANTE's performance against several strong baseline models as above. We follow the training, validation, and testing data splits in The Well~\cite{ohana2024thewell}.

Here we present a detailed description of the benchmarks, which may include the physical settings, governing equations, dataset specifics, and relevant parameters. 

\subsubsection{Turbulent Radiative Layer (TR)}
This dataset explores the dynamics of turbulent radiative layers in astrophysical systems, where hot and cold gases mix, leading to the formation of intermediate-temperature gas that rapidly cools. The simulations model the Kelvin-Helmholtz instability in a 2D domain, with cold, dense gas at the bottom and hot, dilute gas at the top:
\begin{align*}
\frac{ \partial \rho}{\partial t} + \nabla \cdot \left( \rho \vec{v} \right) &= 0, \\
\frac{ \partial \rho \vec{v} }{\partial t} + \nabla \cdot \left( \rho \vec{v}\vec{v} + P \right) &= 0, \\
\frac{ \partial E }{\partial t} + \nabla \cdot \left( (E + P) \vec{v} \right) &= - \frac{E}{t_{\rm cool}}, \\
E = P / (\gamma -1) \, \, \gamma &= 5/3,
\end{align*}
where $\rho$ is the density, $\vec{v}$ is the 2D velocity, $P$ is the pressure, $E$ is the total energy, and $t_{\rm cool}$ is the cooling time. The data capture key phenomena such as mass flux from the hot to cold phase, turbulent velocities, and the distribution of mass across temperature bins. The dataset includes $101$ timesteps of $384\times128$ resolution for $90$ trajectories, varying the cooling time \( t_{\rm cool} \) across nine values. Simulations were performed using Athena++ on a uniform Cartesian grid with periodic boundary conditions in the $x$-direction and zero-gradient in the $y$-direction. This dataset provides insights into the phase structure, energetics, and dynamics of multiphase gas in astrophysical environments, such as the interstellar and circumgalactic media.

\subsubsection{Active Matter (AM)}
\label{am_details}
This dataset comprises simulations of a continuum theory describing the dynamics of \(N\) rod-like active particles in a Stokes fluid within a two-dimensional domain of linear size \(L\). The data include $81$ time-steps of \(256 \times 256\) resolution per trajectory, with fields such as concentration (scalar), velocity (vector), orientation tensor $D$, and strain-rate tensor $E$. Simulations explore parameter variations in alignment (\(\zeta\)), dipole strength (\(\alpha\)), and other coefficients, capturing phenomena like energy transfer across scales, vorticity-orientation coupling, and the isotropic-to-nematic phase transition. Periodic boundary conditions and uniform Cartesian grids are employed, with data stored at $0.25$-second intervals over a $20$-second timespan. 

Specifically, the particle configuration is described by a continuum distribution function $\Psi(\mathbf{x}, \mathbf{p}, t)$, which describes the center of mass $\mathbf{x}$ and orientation vector $\mathbf{p}$ at time $t$. $\langle \cdot \rangle = \int_{|\mathbf{p}|=1} \cdot \, f \, d\mathbf{p}$ denotes an orientational moment. 
The evolution of $\Psi(\mathbf{x}, \mathbf{p}, t)$ is governed by the Smoluchowski equation, which ensures particle number conservation,
\begin{equation*}
\frac{\partial \Psi}{\partial t} + \nabla_\mathbf{x} \cdot (\mathbf{\dot{x}} \Psi) + \nabla_\mathbf{p} \cdot (\mathbf{\dot{p}} \Psi) = 0,
\end{equation*}
where the conformational fluxes $\mathbf{\dot{x}}$ and $\mathbf{\dot{p}}$ are derived from the dynamics of a single particle within a background flow $\mathbf{u}(\mathbf{x}, t)$. In the case of a concentrated suspension, these fluxes are given by
\begin{equation}
\label{advect}
\mathbf{\dot{x}} = \mathbf{u} - d_T \nabla_\mathbf{x} \log \Psi,
\end{equation}
\begin{equation}
\label{rotate}
\mathbf{\dot{p}} = (I - \mathbf{p} \mathbf{p}) \cdot (\nabla \mathbf{u} + 2\mathbf{D}) \cdot \mathbf{p} - d_R \nabla_\mathbf{p} \log \Psi.
\end{equation}
Here, $d_T$ and $d_R$ represent dimensionless translational and rotational diffusion constants, $\zeta$ is the alignment strength, and $\mathbf{D} = \langle \mathbf{p} \mathbf{p} \rangle$ is the second-moment tensor. Eq.~\eqref{advect} describes the particles being advected by the local mean-field velocity $\mathbf{u}$ while diffusing isotropically. Similarly, Eq.~\eqref{rotate} models the angular flux velocity, determined by both the rotation of slender rodlike particles and rotational diffusion. The Smoluchowski equation is coupled to the Stokes flow as
\begin{equation*}
- \Delta \mathbf{u} + \nabla \cdot P = \nabla \cdot \mathbf{\Sigma}, \quad \nabla \cdot \mathbf{u} = 0,
\end{equation*}
\begin{equation*}
\mathbf{\Sigma}=\alpha\mathbf{D} + \beta\mathbf{S}:\mathbf{E} - 2\zeta\beta(\mathbf{D}\cdot \mathbf{D} - \mathbf{S}:\mathbf{D}).
\end{equation*}
Here, $P(\mathbf{x}, t)$ denotes fluid pressure, $\alpha$ is the dimensionless active dipole strength, $\beta$ represents particle density, $\mathbf{E} = [\nabla \mathbf{u} + \nabla^{\mathrm{T}} \mathbf{u}]$ is the symmetric rate-of-strain tensor, and $\mathbf{S} = \langle \mathbf{p} \mathbf{p} \mathbf{p} \mathbf{p} \rangle$ is the fourth-moment tensor. The stress tensor $\mathbf{\Sigma}$ is composed of contributions from viscous particle drag, particle rigidity, and local steric torques. These equations form a closed system, which is solved iteratively to track the kinetic behavior.

\subsubsection{Viscoelastic Fluids (VF)}
\label{vf_details}
This dataset explores the multistability of viscoelastic fluids in a two-dimensional channel flow, capturing four distinct attractors: the laminar state (LAM), a steady arrowhead regime (SAR), Elasto-inertial turbulence (EIT), and a chaotic arrowhead regime (CAR). These states coexist for the same set of parameters, with their emergence dependent on initial conditions:
\begin{equation*}
Re(\partial_t \mathbf{u^\ast} + (\mathbf{u^\ast}\cdot\nabla)\mathbf{u^\ast} ) + \nabla p^\ast = \beta \Delta \mathbf{u^\ast} + (1-\beta)\nabla\cdot \mathbf{T}(\mathbf{C^\ast}),
\end{equation*}
\begin{equation*}
\partial_t \mathbf{C^\ast} + (\mathbf{u^\ast}\cdot\nabla)\mathbf{C^\ast} +\mathbf{T}(\mathbf{C^\ast}) = \mathbf{C^\ast}\cdot\nabla \mathbf{u^\ast} + (\nabla \mathbf{u^\ast})^T \cdot \mathbf{C^\ast} + \epsilon \Delta \mathbf{C^\ast}, 
\end{equation*}
\begin{equation*}
\nabla \mathbf{u^\ast} = 0,
\end{equation*}
\begin{equation*}
\textrm{with} \quad \mathbf{T}(\mathbf{C^\ast}) = \frac{1}{\text{Wi}}(f(\textrm{tr}(\mathbf{C^\ast}))\mathbf{C^\ast} - \mathbf{I}),\qquad
\textrm{and} \quad f(s) := \left(1- \frac{s-3}{L^2_{max}}\right)^{-1},
\end{equation*}
where $\mathbf{u^\ast} = (u^\ast,v^\ast)$ is the streamwise and wall-normal velocity components, $p^\ast$ is the pressure, $\mathbf{C^\ast}$ is the positive definite conformation tensor which represents the ensemble average of the product of the end-to-end vector of the polymer molecules. In two dimensional space, $4$ components of the tensor are solved: $c_{xx}^\ast, c_{yy}^\ast, c_{zz}^\ast, c_{xy}^\ast$. $\mathbf{T}(\mathbf{C^\ast})$ is the polymer stress tensor given by the FENE-P model.
The dataset includes snapshots of these attractors as well as edge states, which lie on the boundary between basins of attraction and provide insight into transitions between flow regimes. The data were generated using direct numerical simulations of the FENE-P model, solving for velocity, pressure, and the conformation tensor fields. Key phenomena include chaotic dynamics in EIT and CAR, as well as the multistability of the system. The dataset, comprising $260$ trajectories with $512\times512$ resolution, is valuable for studying viscoelastic turbulence and evaluating simulators capable of capturing these complex flow behaviors. Simulations were performed using the Dedalus framework, with parameters set to $Re=1000$, $Wi=50$, $\beta=0.9$, $\epsilon=2\times10^{-6}$, and $L_{max}=70$.

\subsubsection{Rayleigh-Bénard Convection (RB)}
\label{rb_detail}

This dataset comprises simulations of two-dimensional, horizontally periodic Rayleigh-Bénard Convection, capturing the dynamics of fluid motion driven by thermal gradients. The system consists of a fluid layer heated from below and cooled from above, leading to the formation of convective cells and complex flow patterns. The dataset includes $200$ timesteps of $512\times128$ resolution for $1,750$ simulations, varying the Rayleigh number ($10^6$ to $10^{10}$), Prandtl number ($0.1$ to $10$), and initial buoyancy perturbations. Fields such as buoyancy (scalar), pressure (scalar), and velocity (vector) are provided, with periodic boundary conditions horizontally and Dirichlet conditions vertically. The data, generated using the Dedalus framework, offer insights into turbulent eddies, convection cells, and the sensitivity of flow structures to initial conditions. This dataset is valuable for studying thermal convection phenomena and validating numerical models in fluid dynamics. The problem is formulated as
\begin{equation*}
\frac{\partial b}{\partial t} - \kappa \Delta b = -u\nabla b,
\end{equation*}
\begin{equation*}
\frac{\partial u}{\partial t} - \nu \Delta u + \nabla p - b \vec{e}_z = -u \nabla u,
\end{equation*}
with boundary conditions
\begin{equation*}
b(z=0) = Lz,~~~ b(z=Lz) = 0, ~~~
u(z=0) = u(z=Lz) = 0.
\end{equation*}

\subsection{Model details}
\label{model-details}

\paragraph{TANTE.}
TANTE's encoder consists of a 3-layer convolutional neural network (CNN), and so do the decoders, where the patch size of the patch embedding is $8\times 8$. Details about the Transformer Processor are shown in Table~\ref{variant_config}. We use $n$ 2-layer MLPs to determine $\tilde{r}_t$ from the output of the Transformer Processor. The active function in the model architecture is Gaussian Error Linear Units (GELU)~\cite{hendrycks2023gaussianerrorlinearunits}. The hyperparameters $\varepsilon$ and $m$ are set to $0.5$ and $2$ respectively.

We present a strategy to ensure TANTE variants have comparable parameter count at each level of model size in Section~\ref{scalability}, and the detailed parameter counts are summarized in Table~\ref{param_size}. This shows the consistency of parameter count across TANTE variants at the same size level.

\begin{table}[htbp]
\centering
\caption{Parameter sizes for TANTE variants at small, medium, and large size level.}
\label{param_size}
\begin{tabular}{c c c c }
\toprule
\textbf{\# Params} & \textbf{Small} & \textbf{Medium} & \textbf{Large} \\
\midrule
TANTE-1 & 4.27M & 18.95M & 39.83M \\
TANTE-2 & 4.55M & 19.33M & 40.49M \\
TANTE-3 & 4.83M & 19.70M & 41.16M \\
\bottomrule
\end{tabular}
\end{table}

\paragraph{FNO.}
The Fourier Neural Operator (FNO) applies a Fast Fourier Transform (FFT) along spatial dimensions. Then it multiplies learned complex-valued weights on a limited number of Fourier modes to perform global convolutions in the frequency domain. We implement a 2-D FNO with the width of $48$, using $20$ Fourier modes per dimension across $4$ spectral layers. 

\paragraph{TFNO.} We implement a Tensorized Fourier Neural Operator (TFNO) with the width of $48$, using $20$ Fourier modes per dimension across $4$ tensorized spectral layers. 

\paragraph{AFNO.} We implement an Adaptive Fourier Neural Operator with $8$ AFNO mixer blocks, where the embedding dimension is $256$. The patch size of AFNO is set to $8\times8$ to align with TANTE.

\paragraph{UNO.} We implement a U-Shaped Neural Operator with $5$ Fourier–UNet encoder–decoder blocks. Starting from $38$ feature channels, each encoder block uplifts the feature channels with a factor of $2$ and the decoder blocks reverse this process with the same factor. Skip connections are adopted at each encoder-decoder level.

\paragraph{CNextUNet.} We implement a  ConvNeXt U-Net with $4$ encoder-decoder stages, each has $4$ ConvNeXt blocks. The encoders start with $15$ initial features and uplift the feature channels with a factor of $2$. The decoders mirror the encoders with transposed CNNs and skip concatenations.

\paragraph{AttUNet.} We implement the Attention U-Net with $5$ encoder-decoder blocks, each employs attention gates in the skip paths. The encoders start with $64$ feature channels and double the number of channels at each level. The decoders mirror the encoders and employ attention gates with the skip connections.

\paragraph{DPOT.} We adopt Denoising Pre-training Operator Transformer (DPOT) at three levels of parameter size according to the original paper~\cite{hao2024dpot}. All variants share AFNO mixers, GELU activations and a patch size of $32{\times}32$. Table~\ref{dpot_variant} summarizes details about the model configurations. 

\begin{table}[htbp]
\centering
\caption{Configurations of DPOT at different size levels.}
\label{dpot_variant}
\begin{tabular}{c c c c c c}
\toprule
\textbf{Size} & \textbf{Embed dim.} & \textbf{MLP dim.} & \textbf{\# Heads} & \textbf{\# Blocks} & \textbf{\# Params} \\ \midrule
Small & 1024 & 1024 & 8 & 6 & 32M \\ 
Medium & 1024 & 4096 &  8 & 12 & 112M \\ 
Large & 1536 & 6144 &  16 & 24 & 540M \\ \bottomrule
\end{tabular}
\end{table}

\paragraph{AViT.} We adopt Axial Vision Transformer following the original paper~\cite{NEURIPS2024_d7cb9db5}. The AViT has $12$ space-time blocks with $6$ attention heads, where the embedding dimension is $384$.

\paragraph{CViT.} We implement CViT using the PyTorch deep learning framework to ensure compatibility with our codebase, as the original CViT model was developed using the JAX framework. CViT has $10$ encoder layers with $8$ attention heads, where the embedding dimension is $512$ and the MLP ratio is $1\times$. We use a patch size of $8\times8$ that is comparable with TANTE. We adopt the grid-based coordinate embeddings in CViT which show the best performance in the original paper~\cite{wang2025cvitcontinuousvisiontransformer}. 

CViT employs a training strategy that randomly samples $Q=2^{10}$ query coordinates from the input grid and corresponding output labels within each batch. This approach serves as a trade-off to mitigate the significant computational overhead introduced by the grid coordinate embedding for cross-attention. 
For completeness, we use the full set of coordinates for training on the \textit{TR} and \textit{AM} datasets to query the latent space. However, using the full coordinates for training is not feasible on the \textit{VF} and \textit{RB} datasets due to their high spatial resolution. After observing that the model performance on the \textit{VF} dataset remains unaffected when $Q$ ranges from $2^{10}$ to $2^{14}$, we adopt the original random sampling for training on these two benchmarks. Interestingly, we did not observe any training instabilities discussed in this paper when clipping the loss gradients to a maximum norm of $1$. 

\subsection{Baseline selection}
\label{baseline-choice}
Our selection of baselines is guided by the principle of ensuring fair and meaningful comparisons within our specific problem setting. Many existing operator learning methods were not included for the following reasons.

\paragraph{Outdated or niche methods.}
We focused on more representative and widely recognized models. Earlier approaches (e.g., DeepONet, graph neural operator, early Galerkin Transformer or operator Transformer variants) have been largely superseded or lack broad adoption as standard benchmarks, thus offering limited new insights.

\paragraph{Superseded approaches.}
We prioritized comparison against the latest and best-in-class models that have demonstrably outperformed their predecessors. For instance, we consider CViT and AViT as the leading ViT-based operator learning models, and DPOT as a representative large transformer-based neural operator, ensuring evaluation against the strongest contemporary competitors.

\paragraph{Incompatible baselines.}
Several notable recent models are not suitable for our setting due to fundamental differences in their problem assumptions or target applications. For example, Physics-Informed Neural Operator requires prior knowledge of PDE equations and boundary conditions, which is not assumed in our benchmarks. Other specialized frameworks (e.g., Transolver, universal physics Transformer, general neural operator Transformer, and geometry-informed neural operator) are designed for PDEs on complex geometries or irregular meshes, which are irrelevant to our regular-grid benchmarks.

\subsection{Computational cost}
\label{computational-cost}

Table~\ref{training_times} reports the training time for all TANTE and baselines on the Turbulent Radiative Layer benchmark. All experiments are conducted on a single NVIDIA H100 GPU. We observe that the Neural Taylor Expansion does not significantly increase training time compared to other models. The overhead from predicting higher-order terms and the penalty of radius $\tilde{r}_t$ is modest.

\begin{table}[htbp]
\centering
\caption{Training times of the Turbulent Radiative Layer benchmark (100,000 iterations) on a single NVIDIA H100 GPU.}
\label{training_times}
\setlength{\tabcolsep}{5pt}
\begin{minipage}[t]{0.48\linewidth}
\centering
\begin{tabular}{l c c}
\toprule
\textbf{Model} & \textbf{\# Params} & \textbf{Time (min)} \\
\midrule
FNO & 4.1M & 48.7 \\
TFNO & 4.1M & 48.7 \\
AFNO & 4.9M & 59.8 \\
UNO & 30.4M & 236.9 \\
CNextUNet & 4.0M & 92.1 \\ 
AttUNet & 34.9M & 87.7 \\ 
DPOT-S & 32.1M & 179.3 \\
\bottomrule
\end{tabular}
\end{minipage}
\hfill
\begin{minipage}[t]{0.48\linewidth}
\centering
\begin{tabular}{lcc}
\toprule
\textbf{Model} & \textbf{\# Params} & \textbf{Time (min)} \\ 
\midrule
DPOT-M & 112.0M & 196.5 \\ 
DPOT-L & 539.8M & 384.0 \\
AViT & 29.0M & 109.4 \\ 
CViT & 30.9M & 274.8 \\
TANTE-0 & 4.4M & 76.5 \\ 
TANTE-1 & 4.5M & 84.5 \\
TANTE-2 & 4.8M & 106.5 \\
\bottomrule
\end{tabular}
\end{minipage}
\end{table}

\subsection{Additional results}
\label{additional-results}

The main results with relative $L^2$ error are reported in Section~\ref{main-results}. For completeness, Tables~\ref{mse_4} and~\ref{mse_8} provide the corresponding results with the MSE metric.

\begin{table}[htbp]
\centering
\caption{MSE (↓ lower is better) of rollouts with standard deviations on 4 time steps on four benchmarks: Turbulent Radiative Layer (TR), Active Matter (AM), Viscoelastic Fluids (VF), and Rayleigh-Bénard Convection (RB). The best, second-best and third-best results are shown in \textbf{bold} and awarded \Gold, \Silver \ and \Bronze \ separately. We mark our TANTE models with \textcolor{red!68!black}{red font}.}
\label{mse_4}
\resizebox{\textwidth}{!}{
\begin{tabular}{l c
                c@{\hspace{1pt}}c
                c@{\hspace{1pt}}c
                c@{\hspace{1pt}}c
                c@{\hspace{1pt}}c}
\toprule
\textbf{Model} & \textbf{\# Params} & \textbf{TR} & & \textbf{AM} & & \textbf{VF} & & \textbf{RB} & \\
\midrule
FNO      & 4M   & 0.2481 $\pm$ 0.1081 & & 0.3569 $\pm$ 0.0922 & & 0.0918 $\pm$ 0.0039 & & 0.1694 $\pm$ 0.0172 & \\
TFNO     & 4M   & 0.2480 $\pm$ 0.1076 & & 0.3556 $\pm$ 0.0918 & & 0.0924 $\pm$ 0.0039 & & 0.1691 $\pm$ 0.0172 & \\
AFNO     & 5M   & 0.1652 $\pm$ 0.0535 & & 0.1293 $\pm$ 0.0136 & & 0.1293 $\pm$ 0.0042 & & 0.0731 $\pm$ 0.0051 & \\
UNO      & 30M  & 0.1932 $\pm$ 0.0728 & & 0.1234 $\pm$ 0.0135 & & 0.1281 $\pm$ 0.0007 & & 0.3913 $\pm$ 0.0384 & \\
\noalign{\vskip 2pt}
CNextUNet& 4M   & 0.1555 $\pm$ 0.0451 & & 0.1369 $\pm$ 0.0144 & & 0.0814 $\pm$ 0.0034 & & 0.0934 $\pm$ 0.0064 & \\
AttUNet  & 35M  & 0.1691 $\pm$ 0.0629 & & 0.1374 $\pm$ 0.0237 & & 0.0377 $\pm$ 0.0006 & & 0.0707 $\pm$ 0.0051 & \\
\noalign{\vskip 2pt}
DPOT-S   & 32M  & 0.1865 $\pm$ 0.0647 & & 0.2279 $\pm$ 0.0296 & & 0.1221 $\pm$ 0.0041 & & 0.0681 $\pm$ 0.0036 & \\
DPOT-M   & 112M & 0.1694 $\pm$ 0.0570 & & 0.2127 $\pm$ 0.0247 & & 0.1100 $\pm$ 0.0039 & & 0.0772 $\pm$ 0.0060 & \\
DPOT-L   & 540M & 0.1726 $\pm$ 0.0547 & & 0.2326 $\pm$ 0.0245 & & 0.0907 $\pm$ 0.0026 & & 0.0902 $\pm$ 0.0077 & \\
\noalign{\vskip 2pt}
AViT     & 29M  & 0.1883 $\pm$ 0.0690 & & 0.1724 $\pm$ 0.0254 & & 0.0205 $\pm$ 0.0004 & & 0.0861 $\pm$ 0.0068 & \\
CViT     & 31M  & \textbf{0.1420 $\pm$ 0.0520} & \Bronze & \textbf{0.0439 $\pm$ 0.0027} & \Bronze & 0.1074 $\pm$ 0.0045 & & 0.2709 $\pm$ 0.0286 & \\
\hline
\noalign{\vskip 3pt}
\textcolor{red!68!black}{TANTE-0}   & 4.41M   & 0.1675 $\pm$ 0.0691 & & 0.0532 $\pm$ 0.0035 & & \textbf{0.0143 $\pm$ 0.0012} & \Bronze & \textbf{0.0310 $\pm$ 0.0017} & \Bronze \\
\textcolor{red!68!black}{TANTE-1}   & 4.52M   & \textbf{0.1245 $\pm$ 0.0411} & \Silver & \textbf{0.0360 $\pm$ 0.0028} & \Gold & \textbf{0.0125 $\pm$ 0.0010} & \Silver & \textbf{0.0261 $\pm$ 0.0012} & \Gold \\
\textcolor{red!68!black}{TANTE-2}   & 4.81M   & \textbf{0.1064 $\pm$ 0.0313} & \Gold & \textbf{0.0366 $\pm$ 0.0027} & \Silver & \textbf{0.0123 $\pm$ 0.0010} & \Gold & \textbf{0.0306 $\pm$ 0.0016} & \Silver \\
\bottomrule
\end{tabular}
}
\end{table}

\begin{table}[htbp]
\centering
\caption{MSE (↓ lower is better) of rollouts with standard deviations on 8 time steps on four benchmarks: Turbulent Radiative Layer (TR), Active Matter (AM), Viscoelastic Fluids (VF), and Rayleigh-Bénard Convection (RB). The best, second-best and third-best results are shown in \textbf{bold} and awarded \Gold, \Silver \ and \Bronze \ separately. We mark our TANTE models with \textcolor{red!68!black}{red font}.}
\label{mse_8}
\resizebox{\textwidth}{!}{
\begin{tabular}{l c
                c@{\hspace{1pt}}c
                c@{\hspace{1pt}}c
                c@{\hspace{1pt}}c
                c@{\hspace{1pt}}c}
\toprule
\textbf{Model} & \textbf{\# Params} & \textbf{TR} & & \textbf{AM} & & \textbf{VF} & & \textbf{RB} & \\
\midrule
FNO      & 4M   & 0.2830 $\pm$ 0.1292 & & 0.5566 $\pm$ 0.1495 & & 0.0878 $\pm$ 0.0038 & & 0.2407 $\pm$ 0.0267 & \\
TFNO     & 4M   & 0.2832 $\pm$ 0.1281 & & 0.5535 $\pm$ 0.1478 & & 0.0886 $\pm$ 0.0038 & & 0.2407 $\pm$ 0.0326 & \\
AFNO     & 5M   & 0.2178 $\pm$ 0.0849 & & 0.2880 $\pm$ 0.0573 & & 0.1265 $\pm$ 0.0039 & & 0.1105 $\pm$ 0.0090 & \\
UNO      & 30M  & 0.2298 $\pm$ 0.0959 & & 0.2872 $\pm$ 0.0664 & & 0.1453 $\pm$ 0.0012 & & 0.6345 $\pm$ 0.1212 & \\
\noalign{\vskip 2pt}
CNextUNet& 4M   & 0.1928 $\pm$ 0.0636 & & 0.3307 $\pm$ 0.0710 & & 0.0803 $\pm$ 0.0036 & & 0.1432 $\pm$ 0.0121 & \\
AttUNet  & 35M  & 0.2233 $\pm$ 0.0971 & & 0.3667 $\pm$ 0.1068 & & 0.0671 $\pm$ 0.0020 & & 0.1301 $\pm$ 0.0126 & \\
\noalign{\vskip 2pt}
DPOT-S   & 32M  & 0.2230 $\pm$ 0.0874 & & 0.4358 $\pm$ 0.0955 & & 0.1193 $\pm$ 0.0034 & & 0.1004 $\pm$ 0.0071 & \\
DPOT-M   & 112M & 0.2037 $\pm$ 0.0761 & & 0.3857 $\pm$ 0.0862 & & 0.1072 $\pm$ 0.0038 & & 0.1310 $\pm$ 0.0142 & \\
DPOT-L   & 540M & 0.2095 $\pm$ 0.0758 & & 0.3795 $\pm$ 0.0721 & & 0.0980 $\pm$ 0.0033 & & 0.1548 $\pm$ 0.0190 & \\
\noalign{\vskip 2pt}
AViT     & 29M  & 0.2490 $\pm$ 0.1130 & & 0.3978 $\pm$ 0.1178 & & 0.0375 $\pm$ 0.0009 & & 0.1366 $\pm$ 0.0145 & \\
CViT     & 31M  & \textbf{0.1847 $\pm$ 0.0801} & \Bronze & 0.1287 $\pm$ 0.0254 & & 0.0976 $\pm$ 0.5093 & & 0.2955 $\pm$ 0.0308 & \\
\hline
\noalign{\vskip 3pt}
\textcolor{red!68!black}{TANTE-0}   & 4.41M   & 0.2017 $\pm$ 0.0884 & & \textbf{0.1147 $\pm$ 0.0165} & \Bronze & \textbf{0.0235 $\pm$ 0.0020} & \Bronze & \textbf{0.0651 $\pm$ 0.0053} & \Bronze \\
\textcolor{red!68!black}{TANTE-1}   & 4.52M   & \textbf{0.1701 $\pm$ 0.0676} & \Silver & \textbf{0.0946 $\pm$ 0.0171} & \Gold & \textbf{0.0208 $\pm$ 0.0017} & \Silver & \textbf{0.0537 $\pm$ 0.0041} & \Gold \\
\textcolor{red!68!black}{TANTE-2}   & 4.81M   & \textbf{0.1555 $\pm$ 0.0598} & \Gold & \textbf{0.0984 $\pm$ 0.0189} & \Silver & \textbf{0.0186 $\pm$ 0.0018} & \Gold & \textbf{0.0549 $\pm$ 0.0050} & \Silver \\
\bottomrule
\end{tabular}
}
\end{table}

\subsection{Mann-Whitney U test}
\label{u-test}

The Mann-Whitney \textit{U} test is a non-parametric method for examining whether two independent samples originate from the same continuous distribution without presuming normality. Given two samples with sizes $n_{1}$ and $n_{2}$, let $R_{1}$ and $R_{2}$ denote the sums of the ranks assigned after pooling and ordering all observations.  
The test statistics are
\begin{equation*}
U_{1}=n_{1}n_{2}+\frac{n_{1}(n_{1}+1)}{2}-R_{1},
\ \
U_{2}=n_{1}n_{2}+\frac{n_{2}(n_{2}+1)}{2}-R_{2},
\end{equation*}
and one sets $U=\min(U_{1},U_{2})$.  
Under the null hypothesis $H_{0}$ (identical distributions), the sampling distribution of $U$ is known exactly for small samples. For $n_{1},n_{2}\ge 20$, it is well-approximated by a normal deviate
\begin{equation*}
Z=\frac{U-\mu_{U}}{\sigma_{U}}, \ \
\mu_{U}=\frac{n_{1}n_{2}}{2}, \ \
\sigma_{U}=\sqrt{\frac{n_{1}n_{2}(n_{1}+n_{2}+1)}{12}},
\end{equation*}
from which a two-sided $p$-value is obtained.  

Because it relies only on ranks, the Mann-Whitney \textit{U} test is robust to outliers and suitable when sample distributions are skewed or ordinal in nature, making it ideal for the comparative analyses reported in this study. We therefore apply the Mann-Whitney \textit{U} test in the \emph{Adaptivity Across System Parameters} experiment (Section~\ref{cross-param}), where ample independent trajectories satisfy the test’s asymptotic requirements.  
Conversely, we omit it in the \emph{Temporal Adaptivity Within Trajectories} experiment (Section~\ref{cross-region}), whose limited data of initial stages would yield an under-powered inference.

\bibliographystyle{unsrtnat}
\bibliography{references}  

\end{document}